 \documentclass[final,5p,times,twocolumn]{elsarticle}

\usepackage{amssymb}
\usepackage{latexsym}
\usepackage{graphicx}
\usepackage{epsfig}
\usepackage{amsfonts}
\usepackage{float}
\usepackage{amsmath}
\usepackage{array}
\usepackage{color}
\usepackage{hyperref}
\usepackage{xcolor}
\usepackage{epstopdf}
\usepackage[tight,small]{subfigure}
 \usepackage{multirow}
\usepackage{mathtools}
\usepackage{amsmath}   
\usepackage{rotating}
\usepackage{balance}
\biboptions{sort&compress}

\journal{international journal for review}

\begin{document}

\begin{frontmatter}



\title{Perceived Safety in Physical Human Robot Interaction - A Survey}



\author{Matteo Rubagotti}
\author{Inara Tusseyeva}
\author{Sara Baltabayeva}
\author{Danna Summers}
\author{Anara Sandygulova}


\tnotetext[]{M. Rubagotti, I. Tusseyeva and A. Sandygulova are with the Department of Robotics and Mechatronics, Nazarbayev University. Nur-Sultan, Kazakhstan. S. Baltabayeva and D. Summers are with the Department of Pedagogy and Psychology, S. Baishev University. Aktobe, Kazakhstan. Corresponding author: Anara Sandygulova, contact email address: \texttt{anara.sandygulova@nu.edu.kz}}
\begin{abstract}
This review paper focuses on different aspects of perceived safety for a number of autonomous physical systems. This is a major aspect of robotics research, as more and more applications allow human and autonomous systems to share their space, with crucial implications both on safety and on its perception. The alternative terms used to express related concepts (e.g., psychological safety, trust, comfort, stress, fear, and anxiety) are listed and explained. Then, the available methods to assess perceived safety (i.e., questionnaires, physiological measurements, behavioral assessment, and direct input devices) are described. Six categories of autonomous systems are considered (industrial manipulators, mobile robots, mobile manipulators, humanoid robots, drones, and autonomous vehicles), providing an overview of the main themes related to perceived safety in the specific domain, a description of selected works, and an analysis of how motion and characteristics of the system influence the perception of safety. The survey also discusses experimental duration and location of the reviewed papers as well as identified trends over time.
\end{abstract}



\begin{keyword}
Physical human robot interaction \sep perceived safety \sep trust \sep comfort \sep UAVs \sep self-driving cars.



\end{keyword}

\end{frontmatter}


\section{Introduction}\label{intro}
\subsection{Motivation}
The past decade has seen an increasing number of applications in which autonomous systems shared their physical space with human beings, with well-known examples given by collaborative robots and self-driving cars. Safety plays a paramount role in these applications, as the first requirement of such systems, before any considerations on their performance, is to make sure they do not cause injury to human beings. This has always been a clear concept, already stated in 1942 in the science fiction literature (\lq\lq a robot may not injure a human being or, through inaction, allow a human being to come to harm'' \cite{asimov1942runaround}), years before the first actual robot manipulator was invented by George Devol \cite{moran2007evolution}. A possible definition of safety when interacting with robots is provided by Haddadin and Croft \cite{haddadin2016physical} as \lq\lq ensuring that only mild contusions may occur in worst case scenarios''.  The topic of safety in physical human-robot interaction (pHRI) has been widely studied, and the achieved results have been listed in several survey papers, e.g., \cite{colgate2008safety,pervez2008safe,guiochet2017safety}.

Guaranteeing safety is only the first step towards seamless pHRI. The next important aspect to consider is that a robot must be perceived as safe, in addition to actually being safe. In other words, \lq\lq achieving a positive perception of safety is a key requirement if robots are to be accepted as partners and co-workers in human environments.'' \cite{bartneck2009measurement}.
From a psychology standpoint, the concept of \emph{perceived safety} can refer to \lq\lq very different areas of human life, such as one’s current health status, experienced exposure to crime, financial situation, and social relationships'' \cite{eller2019psychological}. In the field of pHRI, the term describes \lq\lq the user’s perception of the level of danger when interacting with a robot, and the user’s level of comfort during the interaction'' \cite{bartneck2009measurement}. The stress that derives from a lack of perceived safety when continuously interacting with a robot can have negative effects on human health, which are however less evident and more difficult to analyze as compared to a physical trauma cause by a collision with the same robot. This survey paper focuses on perceived safety in pHRI, given the relevance of the topic for the future of robots and autonomous systems in general. In our work, we extend the concept of \emph{robot} to also include relevant autonomous physical systems, i.e., drones and autonomous road vehicles such as self-driving cars.

\subsection{Related survey papers}\label{sec:related}
Perceived safety in pHRI has already been analyzed within different frameworks in a number of review papers. In particular, Bethel et al. (2007) \cite{bethel2007survey} focused on psychophysiologycal measurements (which can also be used to measure perceived safety) in the general field of human-robot interaction (HRI), and studied their use in conjunction with self-report measures, behavioral measures, and task performance, concluding that these methods should be used together to obtain a reliable evaluation of the interaction between human and robot.

Similarly, Bartneck et al. (2009) \cite{bartneck2009measurement} studied how to measure perceived safety in HRI (13 articles were reviewed on this topic), as part of a more general overview that also analyzed the assessment of anthropomorphism, animacy, likeability, and perceived intelligence. The paper introduced the five \emph{Godspeed questionnaires}, aimed at assessing the
users’ perception of robots, including one questionnaire on perceived safety.

In 2017, Lasota et al. \cite{lasota2017survey} surveyed the general field of safe pHRI, including one chapter on perceived safety (namely, \lq\lq safety through consideration of psychological
factors''), in which 25 papers were described. In addition to considering the different assessment methods already mentioned in \cite{bethel2007survey,bartneck2009measurement}, the authors of \cite{lasota2017survey} stated that the adjustment of the robot behavior to achieve perceived safety can be obtained with methods based either on robot features, or on social considerations. Methods in the first group focus on adjustment of the parameters that define the robot motion, i.e., speed, acceleration, distance to the
human, and robot appearance. The main observed limitation of these approaches was that all of these factors interact with each other, which makes it difficult to define guidelines for each single parameter. In the second category, Lasota and coauthors included methods that try to apply social rules (observed in human-human interaction) to pHRI, and that analyze the impact of factors such as personality traits, experience, and culture. For these approaches, the main observed limitation was the difficulty in obtaining this type of information when the system is deployed.

The 2018 survey by Villani et al. \cite{villani2018survey} focused on industrial applications of robots, and in particular on safety and intuitive user interfaces. One subsection, reviewing 12 papers, was dedicated to human factors, considering a broad overview of psychological aspects of pHRI: the main idea was that one should ideally aim at \lq\lq relieving user's cognitive burden when the task to accomplish overloads her/his mental capabilities, adapting the behaviour of the robot and implementing a sufficient level of autonomy'' \cite{villani2018survey}, and a lack of perceived safety would contribute to increasing the mental burden of the operator.

In 2020, the review paper by Zacharaki et al. \cite{zacharaki2020safety} provided a broad overview of safety in pHRI, also including perceived safety, analyzing papers from the point of view of basic robot functions such as perception, cognition
and action. In their section on societal and psychological factors, they analyzed 12 papers, relying on the categorization already described in \cite{lasota2017survey}.

There exist survey papers already published on the actual safety of drones, together with related aspects such as privacy and security (see, e.g., the works published in 2016 by Vattapparamban et al. \cite{vattapparamban2016drones} and Altawy et al. \cite{altawy2016security}), but none of them accounted for perceived safety. As for autonomous road vehicles, there has been a recent surge of interest in the perception of their safety, although, as for drones, the available survey papers are focusing on actual safety rather than on its perception: see, e.g., the 2019 review paper by Mircicu\u{a} et al. \cite{mircicua2019design}.

\subsection{Paper contribution, method, and organization}

The research questions considered in this survey paper are:
\begin{enumerate}
    \item What are the most commonly used terms employed to express presence or absence of perceived safety in pHRI?
    \item What are the employed assessment methods, i.e., how is perceived safety in pHRI measured?
    \item For industrial manipulators, mobile robots, mobile manipulators, humanoid robots, drone and autonomous vehicles, what are the main themes considered in the related papers?
    \item What is, both overall and for each of the above-mentioned robot types, the correlation between robot characteristics and motion, and perceived safety?
    \item What are the main experimental conditions (in terms of location and duration) in the considered works?
\end{enumerate}

In order to determine the list of works to analyze, we reviewed journal papers, conference proceedings, and book chapters published in English language in international venues until the year 2020, with the following characteristics:
\begin{enumerate}
    \item[a)] real-world experiments (either involving an actual robot, a virtual reality setup, or a driving simulator in the case of autonomous vehicles) are described in the paper, with physical interaction (workspace sharing with or without possibility of physical contact) between a moving robot and one or more human participants; 
    \item[b)] the robot motion is either autonomously determined by a motion planning algorithm, or the robot is operated according to the so-called \emph{Wizard-of-Oz} approach \cite{riek2012wizard};
    \item[c)] perceived safety is assessed, either observing the participants' behavior, via measurements of their physiological variables, using questionnaires or direct input devices;
    \item[d)] considerations are made on the connection between the robot behavior and the perception of safety by the human participants.
\end{enumerate}

In order to determine which papers to insert in our survey, we started by analyzing (using the above-listed criteria) those cited in the survey papers \cite{bethel2007survey,bartneck2009measurement,lasota2017survey,villani2018survey,zacharaki2020safety}. We will refer to these works as \emph{initial papers}. As a second step, we analyzed the works cited in the initial papers, together with the works that cited the initial papers (via Google Scholar), which appeared as possible candidates for our search. The same procedure was repeated for one more iteration for the works that were included in the list of suitable papers. Additionally, we ran direct searches on Google Scholar, IEEExplore, ScienceDirect, and Scopus, inserting relevant keywords related to the type of robot (e.g., \emph{industrial manipulator} or \emph{drone}) or to concepts connected to perceived safety (such as \emph{trust} or \emph{stress}). Searching for relevant works has been a major challenge of this survey paper, the reason for it being that, as the field of perceived safety in pHRI is relatively young, there is not unified terminology.

\begin{figure}[tbp]
	\centering
	\includegraphics[width=\columnwidth]{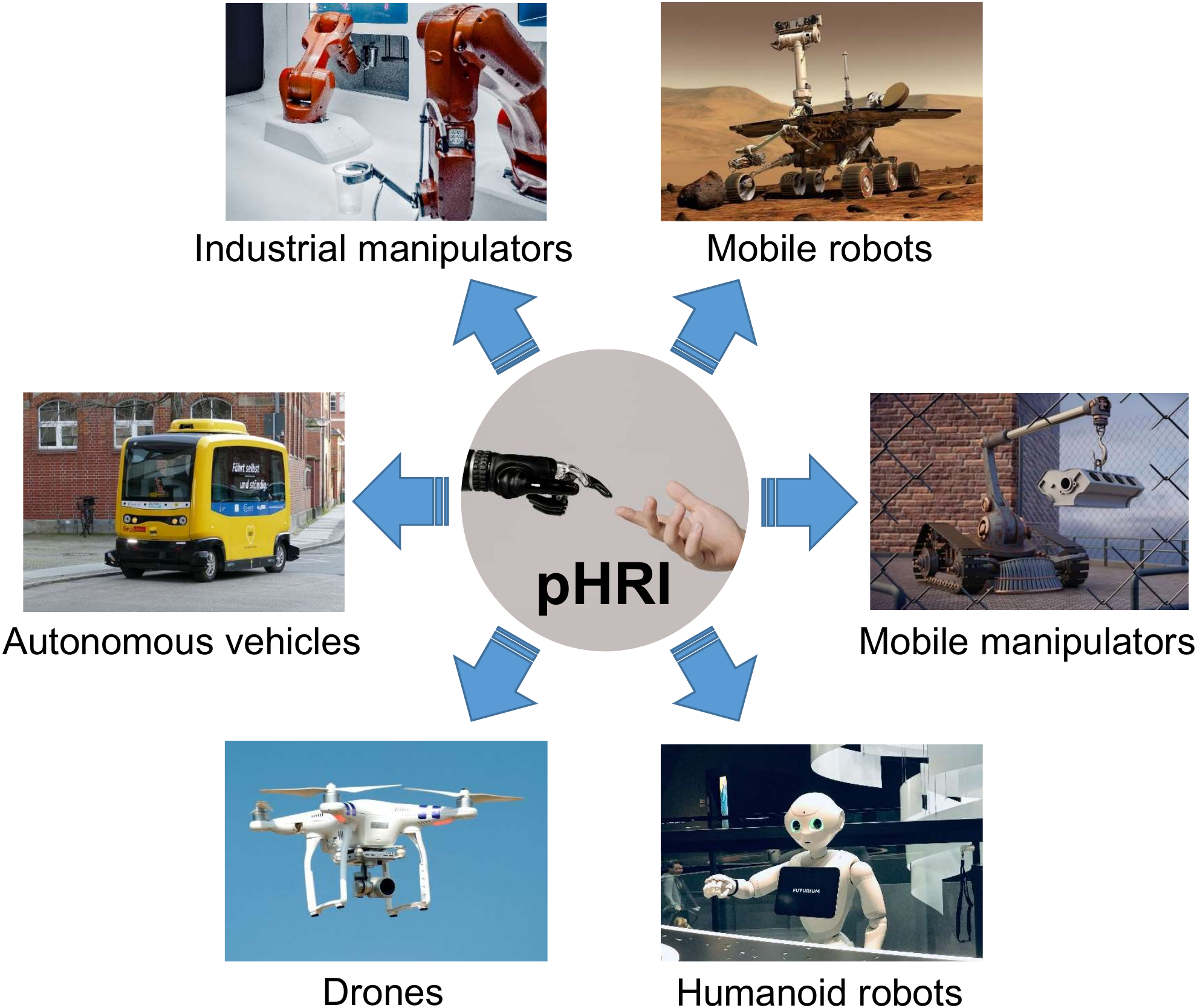}
	\caption{The six types of robots considered in this survey paper.}
	\label{fig:types}
\end{figure}

This survey paper differs from works mentioned is Section~\ref{sec:related} as it analyzes all the listed research questions (while \cite{bethel2007survey,bartneck2009measurement} only focused on assessment methods) and reviews a total of 114 papers entirely focusing on perceived safety in pHRI (while \cite{lasota2017survey,villani2018survey,zacharaki2020safety} were on the broad topic of safety in pHRI, and analyzed a maximum of 25 papers on perceived safety). Also, this survey paper discusses issues related to terminology, and includes drones and autonomous vehicles in its analysis.

The remainder of the paper is organized as follows. In order to provide the needed terminology to the reader, Section \ref{sec:term} will introduce the terms used to describe the idea of perceived safety in the reviewed papers, focusing in particular on psychological/mental/subjective safety, trust, comfort, stress, fear, anxiety and surprise. Section \ref{sec:assess} will review the assessment methods used in all the analyzed works, to measure the participants reactions, attitude and emotions toward the physical interaction with the robot, mainly following the guidelines defined in \cite{bethel2007survey,bartneck2009measurement}, but providing additional considerations. Then, in Sections \ref{sec:IM}-\ref{sec:AVs}, papers will be analyzed based on the type of employed robot (see, Fig. \ref{fig:types}), and a description of the most relevant works will be provided: due to space limitation, we will only describe the most representative papers of the specific category, selected based on number of citations and publication venue. We will analyze how perceived safety in pHRI has been studied for industrial manipulators \cite{yamada1999proposal, hanajima2004influence, hanajima2005motion, kulic2005anxiety, kulic2006estimating, hanajima2006further, kulic2007affective, kulic2007physiological, zoghbi2009evaluation, arai2010assessment, aleotti2012comfortable, ng2012impact, lasota2014toward, weistroffer2014assessing, lasota2015toward, charalambous2016development, rahman2016trust, koppenborg2017effects, maurtua2017human, hocherl2017motion, you2018enhancing, pan2018evaluating, bergman2019close, koert2019learning, wang2019symbiotic, aeraiz2020robot, pollak2020stress, zhao2020task, hu2020interact}, mobile robots \cite{koay2005methodological, koay2006empirical, huttenrauch2006investigating, koay2006methodological, woods2006methodological, syrdal2006doing, dautenhahn2006may, walters2007robotic, walters2008human, syrdal2009negative, karreman2014robot, bhavnani2020attitudes, scheunemann2020warmth}, mobile manipulators \cite{inoue2005comparison, dehais2011physiological, chen2011touched, strabala2013toward, dragan2015effects, brandl2016human, macarthur2017human}, humanoid robots \cite{butler2001psychological, kanda2002development, sakata2004psychological, nomura2004psychology, itoh2006development, koay2007exploratory, edsinger2007human, koay2007living, huber2008human, kanda2008analysis, huber2009evaluation, koay2009five,  takayama2009influences, bainbridge2011robot, de2013relation, chan2013human, moon2014meet, rodriguez2015bellboy, sorostinean2015reliable, haring2015changes, sadrfaridpour2017collaborative, sarkar2017effects, munzer2017impact, ivaldi2017towards, willemse2017affective, neggers2018comfortable, charrier2018empathy, stark2018personal, rajamohan2019factors, block2019softness, fitter2019does, busch2019evaluation, fitter2020exercising, dufour2020visual}, drones \cite{duncan2013comfortable, szafir2014communication, cauchard2015drone, jones2016elevating, duncan2017effects, yeh2017exploring, acharya2017investigation, karjalainen2017social, chang2017spiders, abtahi2017drone, jane2017drone, colley2017investigating, kong2018effects, jensen2018knowing, yao2019autonomous, wojciechowska2019collocated}, and self-driving cars \cite{waytz2014mind, gold2015trust, rothenbucher2016ghost, clamann2017evaluation, eden2017expectation, forster2017increasing, hauslschmid2017supportingtrust, palmeiro2018interaction, reig2018field, rahmati2019influence, ackermann2019experimental, dey2019pedestrian, jayaraman2019pedestrian, nordhoff2020passenger, paddeu2020passenger}. Section~\ref{sec:disc} will report general considerations on factors determining perceived safety, experimental duration and location, and trends on published papers on this topic over time. Conclusions will be finally drawn in Section~\ref{sec:concl}.

\section{Defining perceived safety}\label{sec:term}
The concept expressed in the title and introduction section of this work as \emph{perceived safety} is described in different works using several terms, which are either synonyms, or terms that refer to the same idea from a different angle, or terms that relate to lack of perceived safety, again from different perspectives. In the remainder of this section, we provide the definitions of these terms first from a broad psychological perspective, and then narrowing them down to the field of pHRI.

As the terms listed in Section \ref{sec:syn} are expressing the same concept, they can be used interchangeably without any problems related to their interpretation. Instead, the terms listed in Sections \ref{sec:concepts} and \ref{sec:concepts2} refer to different aspects of perceived safety. Thus, we decided to list, in Table 1, which of these terms are employed in the papers related to each robot type. We can observe that nearly all terms are widely used for each robot type, with the term \emph{comfort} being the most widespread.

\begin{table*}[tbp]
\label{tab:focus}
\centering
\begin{tabular}{|p{0.10\linewidth}|p{0.11\linewidth}|p{0.11\linewidth}|p{0.11\linewidth}|p{0.11\linewidth}|p{0.14\linewidth}|p{0.14\linewidth}|}
\hline
                   & \textbf{Industrial\newline manipulators} & \textbf{Mobile\newline robots} & \textbf{Mobile\newline manipulators} & \textbf{Humanoid robots} & \textbf{Drones} & \textbf{Autonomous\newline vehicles}  \\
\hline\hline
\centering{\textbf{Trust}}         &   \cite{lasota2015toward} \cite{charalambous2016development} \cite{rahman2016trust} \cite{maurtua2017human} \cite{hocherl2017motion}  \cite{you2018enhancing} \cite{bergman2019close} \cite{koert2019learning} \cite{aeraiz2020robot} \cite{hu2020interact}                   &   -                   &  \cite{dragan2015effects} \cite{macarthur2017human}   &    \cite{sadrfaridpour2017collaborative} \cite{sarkar2017effects} \cite{willemse2017affective} \cite{charrier2018empathy}
\cite{block2019softness} \cite{fitter2019does}
\cite{busch2019evaluation} 
\cite{fitter2020exercising}                 &   \cite{cauchard2015drone} \cite{jane2017drone}                 &   \cite{rothenbucher2016ghost} \cite{eden2017expectation} \cite{forster2017increasing}
\cite{reig2018field} \cite{jayaraman2019pedestrian}
\cite{nordhoff2020passenger}                     \\
\hline
\centering{\textbf{Comfort}}         &  \cite{zoghbi2009evaluation} \cite{arai2010assessment} \cite{aleotti2012comfortable} \cite{ng2012impact} \cite{lasota2014toward} \cite{weistroffer2014assessing} \cite{lasota2015toward} \cite{charalambous2016development} \cite{hocherl2017motion} \cite{you2018enhancing} \cite{pan2018evaluating}   \cite{bergman2019close} \cite{koert2019learning} \cite{aeraiz2020robot}   \cite{hu2020interact}                   &  \cite{koay2005methodological} \cite{koay2006empirical} \cite{woods2006methodological} \cite{syrdal2006doing}  \cite{dautenhahn2006may}       \cite{walters2007robotic} \cite{walters2008human} \cite{syrdal2009negative} \cite{karreman2014robot} \cite{scheunemann2020warmth}              &  \cite{inoue2005comparison}
\cite{dehais2011physiological} \cite{chen2011touched}
\cite{strabala2013toward}
\cite{dragan2015effects}
\cite{brandl2016human}                    &   \cite{butler2001psychological} \cite{sakata2004psychological} \cite{koay2007exploratory} \cite{edsinger2007human} \cite{koay2007living} \cite{huber2009evaluation} \cite{koay2009five} \cite{takayama2009influences}   \cite{chan2013human} \cite{rodriguez2015bellboy}  \cite{sadrfaridpour2017collaborative} \cite{sarkar2017effects} \cite{munzer2017impact} \cite{neggers2018comfortable} \cite{charrier2018empathy} \cite{rajamohan2019factors} \cite{block2019softness} \cite{fitter2019does} \cite{busch2019evaluation}   \cite{fitter2020exercising} \cite{dufour2020visual}                  &  \cite{duncan2013comfortable}  \cite{cauchard2015drone}  \cite{jones2016elevating}  \cite{yeh2017exploring} \cite{acharya2017investigation} \cite{karjalainen2017social} \cite{chang2017spiders} \cite{jane2017drone} \cite{kong2018effects}  \cite{jensen2018knowing} \cite{wojciechowska2019collocated}                   &  \cite{rothenbucher2016ghost} \cite{eden2017expectation} \cite{reig2018field} \cite{rahmati2019influence} \cite{ackermann2019experimental} \cite{jayaraman2019pedestrian}  \cite{nordhoff2020passenger}     \cite{paddeu2020passenger}                   \\
\hline\centering{\textbf{Stress}}         & \cite{hanajima2004influence}
\cite{arai2010assessment}
\cite{lasota2014toward}
\cite{lasota2015toward}
\cite{wang2019symbiotic}
\cite{pollak2020stress}   &    \cite{syrdal2006doing}     & \cite{dehais2011physiological}   &  \cite{itoh2006development}
\cite{sorostinean2015reliable} \cite{willemse2017affective}
\cite{dufour2020visual}      &  \cite{yeh2017exploring} \cite{wojciechowska2019collocated}  &  -                                      \\
\hline
\centering{\textbf{Fear}}    & \cite{yamada1999proposal}
\cite{hanajima2006further} 
\cite{arai2010assessment} \cite{you2018enhancing}
\cite{bergman2019close}
&   \cite{walters2007robotic} \cite{walters2008human} \cite{syrdal2009negative}                   &    \cite{chen2011touched} \cite{macarthur2017human}                &   \cite{sakata2004psychological}
\cite{nomura2004psychology}
\cite{bainbridge2011robot}
\cite{sorostinean2015reliable}
\cite{ivaldi2017towards} \cite{munzer2017impact}
\cite{block2019softness}
\cite{fitter2019does}
\cite{fitter2020exercising}                  &  -                   &   \cite{nordhoff2020passenger}                    \\
\hline
\centering{\textbf{Anxiety}}   &  \cite{hanajima2005motion} \cite{kulic2005anxiety}
\cite{hanajima2006further}
\cite{koppenborg2017effects}
\cite{aeraiz2020robot}
\cite{zhao2020task}
   &    \cite{syrdal2009negative}     &  \cite{inoue2005comparison}  &  \cite{sakata2004psychological} \cite{nomura2004psychology} \cite{itoh2006development}      & \cite{yao2019autonomous}     & \cite{paddeu2020passenger}                      \\
\hline
\centering{\textbf{Surprise}}  &   \cite{kulic2005anxiety}
\cite{arai2010assessment}                 &     \cite{scheunemann2020warmth}                 &  \cite{inoue2005comparison}
\cite{chen2011touched}
\cite{dragan2015effects} &  \cite{sakata2004psychological}
\cite{sorostinean2015reliable}
\cite{munzer2017impact}         &    -                 &   -                    \\
\hline
\end{tabular}
\caption{Focus on different aspects of perceived safety by robot type. In each row, the table lists the considered focus of the paper, which can be trust, comfort, stress, fear, anxiety, and/or surprise. In each column, a different robot type is considered, and precisely industrial manipulators, mobile robots, mobile manipulators, humanoid robots, drones, and autonomous vehicles.}
\end{table*}

\subsection{Synonyms of perceived safety}\label{sec:syn}
The most commonly used terms that refer to the same idea as perceived safety (and can thus be considered synonyms) are listed in the following.

\textbf{Psychological safety}. This concept, in the general field of psychology, was described by Edmondson et al. as \lq\lq people’s perceptions of the  consequences of taking interpersonal risks in a particular context such as a workplace'' \cite{edmondson2014psychological}. As stated by Abror and Patrisia \cite{abror2020psychological}, psychological safety could function as a shared belief of a particular group of people, allowing them to take risks. In pHRI, this concept was defined by Lasota et al. as follows: \lq\lq maintaining psychological safety involves ensuring that the human perceives interaction with the robot as safe, and that interaction does not lead to any psychological discomfort or stress as a result of the robot’s motion, appearance, embodiment, gaze, speech, posture, social conduct, or any other attribute'' \cite{lasota2017survey}.

\textbf{Mental safety}. It is sometimes used in psychology as synonym of psychological safety  \cite{schepers2008psychological,pacheco2015silence}. In pHRI, mental safety was defined by Villani et al. as related to the \lq\lq mental stress and anxiety induced by close interaction with robot'' \cite{villani2018survey}, or equivalently, by Sakata et al. \cite{sakata2004psychological}, as the condition in which humans do not feel fear or surprise toward the robot.

\textbf{Subjective safety}. From the general psychological point of view, this term can be described as a general measure reflecting a persons' perception of the security of a particular location, as stated by Patwardhan et al. \cite{patwardhan2020visitors}. Even though we could not find a definition in pHRI papers, such definition was provided for a related framework (the safety of vulnerable road users) by Sorensen and Mosslemi, as \lq\lq the feeling or perception of safety'' \cite{sorensen2009subjective}. In pHRI, this term should not be confused with a feature of personalized safety systems that, as stated by Traver et al., \lq\lq bear in mind the special characteristics of human beings'' \cite{traver2000making}, also sometimes referred to as subjective safety. 

\subsection{Concepts related to perceived safety}\label{sec:concepts}
The following two terms are instead related to perceived safety, although they are often used within a wider range of meanings.

\textbf{Trust}. In psychology, trust is a variable described in a social context as a factor of group formation. Its function is to reduce social complexity and build confidence in the security of the considered \emph{abstract system}, as stated by Mukherjee and Nath \cite{mukherjee2007role}. Also, Ferrin et al. claimed that trust as a belief helps one party rely on another in a situation of \emph{social dilemma} \cite{ferrin2007can}. Although there is no unified definition of trust in pHRI, this concept was broadly defined by Kok and Soh as follows: \lq\lq an agent’s trust in another agent [is defined] as a multidimensional latent variable that mediates the relationship between events in the past and the former agent's subsequent choice of relying on the latter in an uncertain environment'' \cite{kok2020trust}. In our survey, we focus on papers in which trust is intended as follows: \lq\lq how much do the human subjects trust that the robot will not harm them?''. As such, trust is directly related to perceived safety, although not being an exact synonym. It is important to mention that trust can be related to concepts not directly connected to safety, for instance the idea of trusting a robot to complete the assigned task: for example, this interpretation is used by Hancock et al. when claiming that \lq\lq the less an individual trusts a robot, the sooner he or she will intervene as it progresses toward task completion'' \cite{hancock2011meta}.

\textbf{Comfort}. The concept of comfort in psychology is very general: as stated by Pineau, \lq\lq comfort corresponds to everything contributing to the well-being and convenience of the material aspects of life'' \cite{pineau1982psychological}. One often finds the expression \emph{being in the comfort zone}, referring to a state where individuals do not express anxiety, fear, or agitation, as their basic physiological needs are satisfied. 
In pHRI, comfort was defined by Koay et al. as the ability of a robot \lq\lq to perform and provide assistance for certain useful tasks, [behaving] in a socially acceptable manner'' \cite{koay2005methodological}. Similarly to trust, comfort is directly related to perceived safety when by \emph{socially acceptable manner} we intend that the robot motion is not perceived as possibly harmful for humans. For example, Norouzzadeh et al. stated that \lq\lq colliding with a robot [...] would definitely imply the risk of injury which is depicted in the high discomfort rating (negative comfort)'' \cite{norouzzadeh2012towards}. This is the type of comfort (or, conversely, discomfort) in pHRI that is being taken into account in our work.

\subsection{Expressing lack of perceived safety}\label{sec:concepts2}
The most commonly used terms that identify lack of perceived safety are estimated as part of the so-called \emph{affective state} of the human participants, and are stress, fear, anxiety and surprise. These emotional responses can be caused by different factors: in our work we will consider them when deriving from a lack of perceived safety in a pHRI scenario.

\textbf{Stress (or strain)}. 
We refer to the definition of stress provided by Folkman and Lazarus as \lq\lq a particular relationship between the person and the environment that is appraised by the person as taxing or exceeding his or her resources and endangering his or her well-being'' \cite{folkman1984stress}.  In pHRI, stress is determined by \lq\lq the changes in the nature of the work (from physical to mental activities), the proximity of the robot to the human operator and the robot’s movement, or the loss of control that can stem from the automation of robotic agents'', as defined by Pollak et al. in \cite{pollak2020stress}. A related concept is that of \emph{robostress}, defined by Vanni et al. as \lq\lq a human estimated or perceived stress when working with the interactive physical robots'' \cite{vanni2019robostress}. 
In some works, the term \emph{strain} is used with the same meaning.

\textbf {Fear.} 
Fear is an actual emotional response that can impel changes in attitude or behavior intentions, as part of an evolutionary mechanism focused on survival (see, e.g., Perkins et al. \cite{perkins2007fear}). We did not find an explicit definition of fear in the pHRI field, and the term is always used directly: for example, Yamada et al. stated that detecting fear was \lq\lq primarily important for ensuring human emotional security in parallel with human physical safety'' \cite{yamada1999proposal}. 

\textbf{Anxiety.} In psychology, 
anxiety is an emotional state that occurs before some event, and, as stated by Spielberger, ``include[s] feelings of apprehension, tension, nervousness, and worry accompanied by physiological arousal" \cite{spielberger2010state}. In other words, anxiety is an adaptive response that occurs to danger and prepares humans to cope with environmental changes (see, e.g., the work of Guti\'{e}rr\'{e}z-Garc\'{i}a and Contrer \cite{gutierrez2013anxiety}). 
In the field of HRI, Nomura and Kanda defined \emph{robot anxiety} as the \lq\lq  emotions of anxiety or fear preventing individuals from interaction with robots having functions of communication in daily life, in particular, dyad communication with a robot'' \cite{nomura2003proposing}. 

\textbf{Surprise.} In psychology, surprise was defined by Celle et al. as an emotion that \lq\lq emerges when there is a discrepancy between one’s expectations and reality'' \cite{celle2018describing}. 
In pHRI, the term (although not directly defined) can be clearly related to a perceived lack of safety. For example, Arai et al. stated that a robot \lq\lq generates fear and surprise because it looks large and strong enough to harm human body and it moves swiftly and unpredictably enough not to avoid the collision'' \cite{arai2010assessment}, while Norouzzadeh et al.  claimed that \lq\lq being surprised by the reactions of the device causes lower perceived safety'' \cite{norouzzadeh2012towards}.

\subsection{Valence and arousal}
As an alternative to the use of discrete emotion categories, such as happiness, fear and anxiety, a different representation is commonly used in emotion detection research. This is the two-dimensional representation consisting of valence and arousal. Quoting from the work of Kulic and Croft \cite{kulic2005anxiety}, \lq\lq valence measures the degree to which the emotion is positive (or negative), and arousal measures the strength of the emotion''. As compared to using discrete emotion categories, \lq\lq the valence/arousal representation provides less data, but the amount of information retained appears adequate for the purposes of robotic control, and is easier to convert to a measure of user approval'' \cite{kulic2005anxiety}.

\section{Assessment methods}\label{sec:assess}
\begin{figure}[htbp]
	\centering
	\includegraphics[width=\columnwidth]{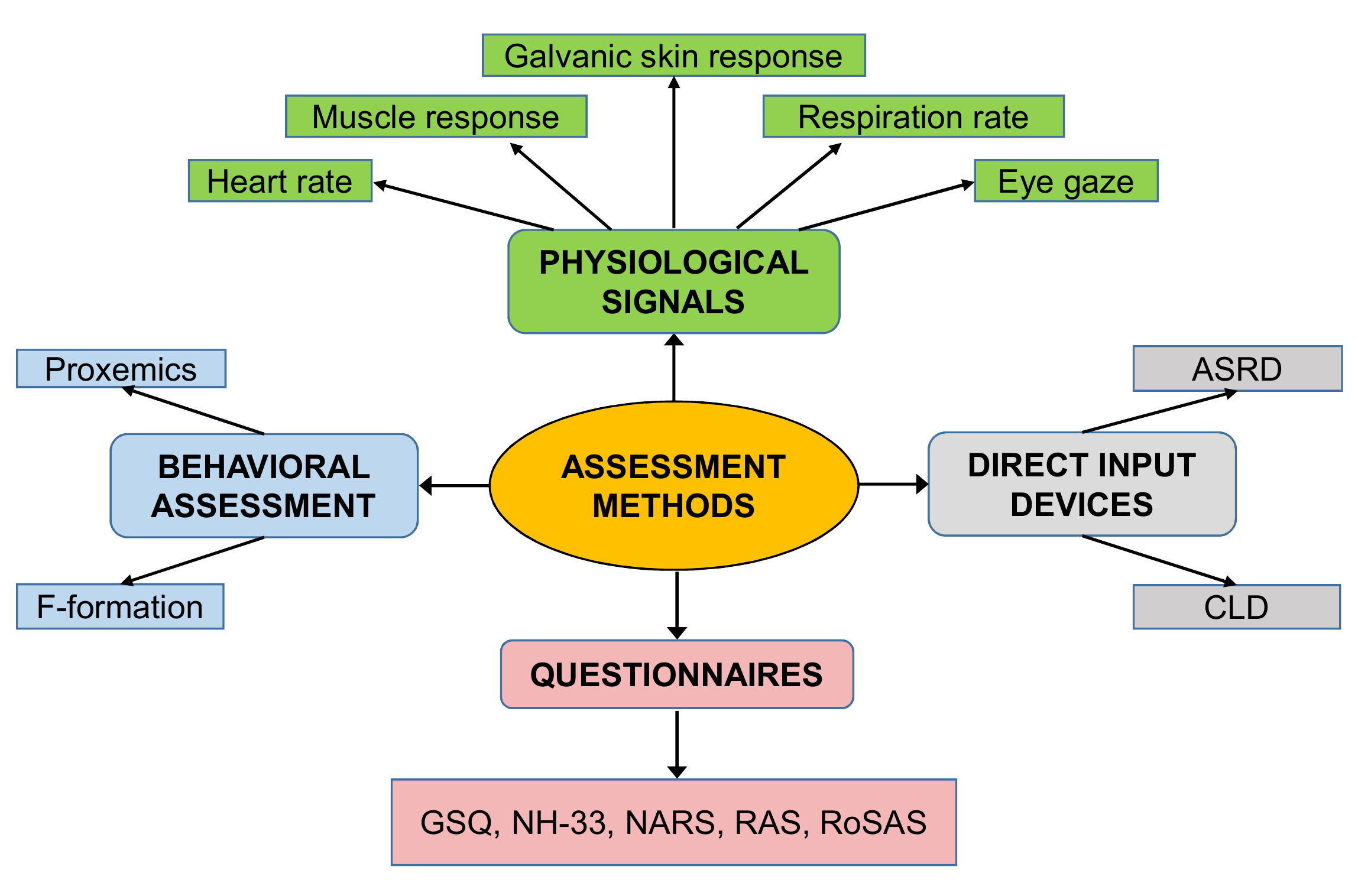}
	\caption{Taxonomy of assessment methods for perceived safety in pHRI.}
	\label{fig:assess}
\end{figure}

In order to measure the level of perceived safety in a pHRI experiment, different methods have been used in the literature, and we divide them in four broad categories: questionnaires, physiological measurements, behavioral assessment, and direct input devices (see the taxonomy in Fig. \ref{fig:assess}). Table 2 lists how the methods in these categories, and their combinations, have been used for different robot types. From the table, one can see that, even if different assessment methods should be used together to obtain a more reliable assessment \cite{bethel2007survey}, several works relied exclusively on questionnaires. This was probably due to the low cost of this method compared, for example, to physiological measurements. Another relatively popular option was to use questionnaires together with behavioural assessment. Physiological measurements were always used in conjunction with questionnaires (probably due again to the low cost of the latter), while few papers used direct input devices.

It is interesting to notice that physiological measurements were never used for mobile robots, drones and autonomous vehicles (with the exception of \cite{jayaraman2019pedestrian}), probably due to the difficulty of using such an assessment method outdoors and/or when the subject did not remain in the same location. Also, the majority of papers considering physiological assessment were on industrial manipulators. A reason for it could be that many experiments involving industrial manipulators were executed with sitting human subjects, a case in which physiological measurements can be acquired more easily and be more reliably. 

\begin{table*}[tbp]
\centering
\begin{tabular}{|p{0.10\linewidth}|p{0.11\linewidth}|p{0.11\linewidth}|p{0.11\linewidth}|p{0.11\linewidth}|p{0.14\linewidth}|p{0.14\linewidth}|}
\hline
                  & \textbf{Industrial\newline manipulators} & \textbf{Mobile\newline robots} & \textbf{Mobile\newline manipulators} & \textbf{Humanoid robots} & \textbf{Drones} & \textbf{Autonomous\newline vehicles}  \\
\hline\hline
\centering{\textbf{Q}}       & \cite{aleotti2012comfortable} \cite{ng2012impact} \cite{lasota2015toward} \cite{charalambous2016development} \cite{rahman2016trust} \cite{hocherl2017motion} \cite{you2018enhancing} \cite{pan2018evaluating}     \cite{bergman2019close} \cite{koert2019learning}      &   \cite{syrdal2006doing} \cite{walters2008human} \cite{syrdal2009negative} \cite{karreman2014robot}                   &  \cite{inoue2005comparison} \cite{strabala2013toward} \cite{dragan2015effects}  \cite{macarthur2017human}                     &  \cite{kanda2002development}  \cite{sakata2004psychological} \cite{nomura2004psychology} \cite{koay2007exploratory}  \cite{huber2009evaluation} \cite{koay2009five} \cite{bainbridge2011robot}  \cite{de2013relation} \cite{chan2013human} \cite{moon2014meet} \cite{sorostinean2015reliable} \cite{sadrfaridpour2017collaborative} \cite{munzer2017impact} \cite{ivaldi2017towards}     \cite{neggers2018comfortable} \cite{charrier2018empathy}     \cite{fitter2019does} \cite{fitter2020exercising} \cite{dufour2020visual}                    & \cite{yeh2017exploring} \cite{karjalainen2017social} \cite{wojciechowska2019collocated}              & \cite{forster2017increasing} \cite{ackermann2019experimental}  \cite{paddeu2020passenger}      \\
\hline
\centering{\textbf{PA}}      &  -   &   -    &   -                &  \cite{itoh2006development} \cite{willemse2017affective}   &   -                    &    -                     \\
\hline
\centering{\textbf{BA}}      &    -                   &    -                   &  -                     &    \cite{edsinger2007human} \cite{rodriguez2015bellboy}                 &  -       &  \cite{rahmati2019influence}                         \\
\hline
\centering{\textbf{Q+PA}}    &  \cite{yamada1999proposal} \cite{hanajima2004influence} \cite{hanajima2005motion} \cite{kulic2005anxiety} \cite{kulic2006estimating} \cite{hanajima2006further}  \cite{kulic2007affective} \cite{kulic2007physiological} \cite{arai2010assessment} \cite{koppenborg2017effects} \cite{aeraiz2020robot} \cite{pollak2020stress} \cite{zhao2020task}                      & -                    &  \cite{dehais2011physiological}                     &   \cite{willemse2017affective}                   &   -                    &     -                    \\
\hline
\centering{\textbf{Q+BA}}    & \cite{maurtua2017human}                      &    \cite{huttenrauch2006investigating} \cite{woods2006methodological} \cite{walters2007robotic} \cite{scheunemann2020warmth}                    &    \cite{brandl2016human}                   &    \cite{butler2001psychological} \cite{kanda2008analysis} \cite{takayama2009influences} \cite{haring2015changes}  \cite{sarkar2017effects} \cite{block2019softness} \cite{rajamohan2019factors}                  &   \cite{duncan2013comfortable} \cite{szafir2014communication} \cite{cauchard2015drone} \cite{duncan2017effects} \cite{abtahi2017drone} \cite{jane2017drone} \cite{colley2017investigating}  &    \cite{rothenbucher2016ghost}                                \\
\hline
\centering{\textbf{Q+PA+BA}} &  \cite{weistroffer2014assessing} \cite{hu2020interact}                     &    -                   &    \cite{chen2011touched}                 &   \cite{stark2018personal} \cite{busch2019evaluation}                    & -                & \cite{jayaraman2019pedestrian}                              \\
\hline
\centering{\textbf{Q+PA+D}}  &  \cite{zoghbi2009evaluation}                     &  -                     &  -                     &  -                    &   -                    &   -                      \\
\hline
\centering{\textbf{Q+BA+D}} &  -                     &   \cite{koay2005methodological} \cite{koay2006empirical} \cite{koay2006methodological}                    &    -                   &    -                  &   -                    &  -    \\       
\hline
\end{tabular}
\label{tab:assessment}
\caption{Assessment methods by robot type. In each row, the table lists the type of assessment, namely questionnaires (Q), physiological assessment (PA), behavioral assessment (BA), direct input devices (D), and their combinations. In each column, a different robot type is considered, and precisely industrial manipulators, mobile robots, mobile manipulators, humanoid robots, drones, and autonomous vehicles.}
\end{table*}


\subsection{Questionnaires}

One of the most commonly used assessment method in pHRI is a \textit{questionnaire} or \textit{survey} - a research instrument or self-report technique to gather data from human participants about different aspects of human-robot interaction. In psychology, questionnaires can be described as ``systems for collecting information from or about people to describe, compare, or explain their knowledge, attitudes, and behavior" \cite{fink2003design}. 

In a typical HRI experiment, participants can be asked to fill a pre-interaction questionnaire administered prior to their intervention with the robot. Such questionnaires can include questions on participants' demographics (e.g. age, gender, height), their prior experience with the robots, personality assessment (e.g. a Big Five Domain Personality Traits Scale \cite{de2000big}, used in \cite{syrdal2006doing, koay2007exploratory, sorostinean2015reliable, ivaldi2017towards}), or any required pre-tests needed (e.g. typing speed or gaming experience for the purposes of counterbalancing). Such questions allow researchers to analyze the relationships between independent variables (e.g. age, gender, height) and dependent variables (e.g. distance). After robot intervention, participants are asked to fill in a post-trial questionnaire to capture their reflections on HRI experience and perceptions of the robot \cite{bainbridge2008effect} and/or a post-test (e.g. to assess learning gains or changes in their perceptions). 

Within the analyzed papers, the following questionnaires were commonly utilized:

\begin{itemize}

\item The \emph{Godspeed Series Questionnaire (GSQ)} was proposed by Bartneck \cite{bartneck2009measurement} and includes five Semantic Differential (SD) \cite{taherdoost2019best} scales relevant to evaluate the perception of the robotic system \cite{weiss2015meta}. The scales intend to measure anthropomorphism, perceived intelligence, likeability, animacy, and perceived safety with 5-point SD items such as ``anxious"-``relaxed",  ``calm"-``agitated”, ``quiescent"-"surprised". GSQ was utilized in \cite{sarkar2017effects, bhavnani2020attitudes, scheunemann2020warmth}. 

\item \emph{NH-33} was firstly presented in  \cite{kamide2012new} with the purpose ``to quantitatively evaluate the degree of safety of specific humanoids" and with the focus on psychological safety of humans. There are 33 7-point Likert scale items on performance, acceptance, harmlessness, humanness, toughness, and agency (e.g. ``this robot does not seem to go out of control", ``this robot does not seem to do injury to a human’s body", etc.). NH-33 was used in \cite{hu2020interact}.

\item \emph{Negative  Attitude  towards  Robots Scale (NARS)} is a questionnaire developed by Nomura et al. \cite{nomura2006measurement} to assess negative attitude towards robots. NARS consists of fourteen 5-point Likert scale items classified into one of the following groups: negative attitudes  toward a) situations  and  interactions with robots (6 items),  b) social  influence  of  robots (5 items),  and  c) emotions  in  interaction  with robots (3 items). NARS was used in \cite{syrdal2009negative, nomura2004psychology, de2013relation, ivaldi2017towards, willemse2017affective, rajamohan2019factors, takayama2009influences}.   

\item \emph{Robot Anxiety Scale (RAS)} is a psychological scale developed by Nomura et al. \cite{nomura2004psychology} with the aim to assess anxiety towards robots. Pilot testing of RAS included people writing open answers on whether or not humans felt anxiety while interacting with the robot. Based on the given answers, some commonly used phrases were identified: ``anxiety toward unpredictability of robots' actions", ``anxiety toward motions or approach of robots" and ``anxiety toward interaction with robots" which they then included in RAS. RAS was used in \cite{de2013relation, ivaldi2017towards, willemse2017affective, rajamohan2019factors}.

\item \emph{Robotic Social Attribute Scale (RoSAS)} was proposed by Carpinella et al. in 2017 \cite{carpinella2017robotic}. It is an 18-item scale to assess perceived social characteristics of robots and how they affect the quality of interaction with robots. RoSAS has three central factors: warmth, competence, and discomfort. The discomfort factor includes the following feelings: aggressive, awful, scary, awkward, dangerous, and strange. The authors suggested that RoSAS had stronger psychometric features compared to GSQ. Among the analyzed articles included in this survey, RoSAS was used in \cite{pan2018evaluating,scheunemann2020warmth, stark2018personal}.

\end{itemize}

\begin{table*}[tbp]
\label{tab:physio}
\centering
\begin{tabular}{|p{0.14\linewidth}|p{0.11\linewidth}|p{0.11\linewidth}|p{0.11\linewidth}|p{0.11\linewidth}|}
\hline
                   & \textbf{Industrial\newline manipulators} &  \textbf{Mobile\newline manipulators} & \textbf{Humanoid robots} & \textbf{Autonomous\newline vehicles}  \\
\hline\hline
\centering{\textbf{HR}}         &   \cite{koppenborg2017effects} \cite{pollak2020stress} \cite{zhao2020task}    &   -                   &    \cite{stark2018personal}  &   -                     \\
\hline
\centering{\textbf{GSR}}         &  \cite{hanajima2004influence} \cite{hanajima2005motion} \cite{hanajima2006further}        \cite{arai2010assessment}  &  \cite{chen2011touched}    &   -          &  -                      \\
\hline\centering{\textbf{EG}}         &   -    & -   &  \cite{busch2019evaluation}   &  \cite{jayaraman2019pedestrian}                                       \\
\hline
\centering{\textbf{MR}}    & \cite{aeraiz2020robot}                     &    -                 &   -               &   -                    \\
\hline
\centering{\textbf{GSR+EG}}  &   \cite{yamada1999proposal}              &     -                 &   -              &   -                    \\
\hline
\centering{\textbf{HR+RR}}   &  - &  -  &  \cite{itoh2006development}   & -                      \\
\hline
\centering{\textbf{HR+GSR}}  &   \cite{weistroffer2014assessing}             &     -             &   -       &   -                    \\
\hline
\centering{\textbf{HR+GSR+EG}}  &   \cite{hu2020interact}                & -          &  -                 &   -                    \\
\hline
\centering{\textbf{HR+GSR+MR}}  &   \cite{kulic2005anxiety} \cite{kulic2006estimating} \cite{kulic2007affective} \cite{kulic2007physiological} \cite{zoghbi2009evaluation}       & -                     &  -         &   -                    \\
\hline
\centering{\textbf{HR+GSR+RR}} & -      &     -                 &  \cite{willemse2017affective}                 &   -                     \\
\hline
\centering{\textbf{GSR+EG+MR}}  &   -     &  \cite{dehais2011physiological}                    &   -                  &  -                     \\
\hline
\end{tabular}
\caption{Physiological assessment by robot type. In each row, the table lists the measured variables, namely heart rate (HR), Galvanic skin response (GSR), eye gaze (EG), muscle response (MR), respiration rate (RR), and their combinations. In each column, a different robot type is considered, and precisely industrial manipulators, mobile manipulators, humanoid robots, and autonomous vehicles (no physiological assessment was conducted for mobile robots and drones).}
\end{table*}

\subsection{Physiological signals}
In the second category, we find methods that involve measurements of physiological signals. This approach, described in detail in \cite{bethel2007survey}, is part of the field of \emph{psychophysiology}, defined by Stem as \lq\lq any research in which the dependent variable (the subject's response) is a physiological measure and the independent variable (the factor manipulated by the experimenter) a behavioral one'' \cite{stern2001psychophysiological}. The use of physiological signals is important in addition to the use of subjective measures such as questionnaires, as the latter may give a limited insight into the involved subconscious and psycho-biological phenomena. In the analyzed papers, the above-mentioned dependent (measured) variables consisted of:
\begin{itemize}
    \item \emph{Heart rate.} The cardiac response, measured by the heart rate, is one of the most important biomarkers related to the activation of the autonomic nervous system, which operates partially independently from the participants, and can provide information about stress and fear (see, e.g. \cite{pollak2020stress, zhao2020task}).
    \item \emph{Galvanic skin response.} Also known as \emph{electrodermal activity}, it provides information on the production levels of sweat glands, related to skin activity, and directly connected with the state of excitation of the sympathetic nerve. Its increased levels are related to the subject's arousal, which can be related to increased levels of stress, fear, anxiety or surprise. Galvanic skin response is made of two components: ``\emph{tonic skin conductance}, the baseline value recorded when no emotional stimulus is applied, and \emph{phasic skin conductance}, the response acquired when environmental and behavioral changes occur'' \cite{bualan2020investigation}. Galvanic skin response is faster than heart rate, but is influenced by muscle contraction, and thus is difficult to use when a collaborative task has to be executed \cite{yamada1999proposal}.
    \item \emph{Eye gaze.} Understanding the subject's gaze pattern is important, as gaze behavior is \lq\lq an indicator for situation awareness'' \cite{gold2015trust}. Typically, when human subjects feel safe, they will be monitoring the possible source of danger (e.g., a robot manipulator) less often. In \cite{yamada1999proposal}, eye gaze was used in conjunction with pupillary dilation. 
    \item \emph{Muscle response.} The most commonly recorded muscle response is that of the corrugator muscle: \lq\lq the corrugator muscle, located just above each eyebrow close to the bridge of the nose, is responsible for the lowering and contraction of the brows, i.e., frowning, which is intuitively associated with negative valence'' \cite{kulic2007physiological}, i.e., negative emotions such as fear and anxiety. Two exceptions are found respectively in \cite{aeraiz2020robot}, in which the response of the participants' biceps was acquired to assess the physiological arousal due to their discomfort, and in \cite{dehais2011physiological}, in which the deltoid muscle activity was measured during a handover task. In all of these cases, muscle activity was measured via electromyography (EMG).
    \item \emph{Respiration rate.} The respiration rate decreases when the subject is relaxed, and increases with arousal \cite{stern2001psychophysiological}. This is related to the fact that, under stress, the same hormones that determine an increase of heart rate also cause a faster respiration, to prepare for a \lq\lq fight of flight'' scenario.
\end{itemize}
Table 3 lists all the analyzed papers in which physiological measurements were present. In the table we can observe that many different combinations of sensors were used in the analyzed papers. As already observed for Table 2, this type of approach has mainly been used for industrial manipulators. The main advantage of using physiological signals as compared to questionnaires is that \lq\lq participants cannot consciously manipulate the activities of their autonomic nervous system'' \cite{bethel2007survey}.

\subsection{Behavioral assessment}
The third type of assessment, employed in papers \cite{sarkar2017effects, maurtua2017human, koay2005methodological, koay2006empirical, huttenrauch2006investigating, koay2006methodological, woods2006methodological, dautenhahn2006may, walters2007robotic, edsinger2007human, kanda2008analysis, haring2015changes, rodriguez2015bellboy, brandl2016human, rajamohan2019factors, block2019softness, busch2019evaluation, scheunemann2020warmth, jayaraman2019pedestrian, rahmati2019influence, rothenbucher2016ghost, duncan2013comfortable, duncan2017effects, abtahi2017drone, cauchard2015drone, hu2020interact}, and mainly used for mobile robots and humanoids, consists in the application of behavioral assessment methods based on video and photo recordings.

A typical measure used in behavioral assessment is the distance that the human keeps from the robot, which is intuitively inversely proportional to the sense of perceived safety. This is related to the concept of \emph{proxemics}, defined by Hall for human-human physical interaction as \lq\lq the  interrelated observations and theories of man’s use of space as a specialized elaboration of culture'' \cite{hall1966hidden}. In particular, Hall took into account four zones used in interpersonal relations, that can be listed from the closest to the farthest as intimate, personal, social, and public zone. In pHRI, proxemics is typically interpreted as the study of human attitude and feelings when the robot enters one of the above-mentioned zones: as a consequence, the relative distance that humans keep from the robot becomes a measurable indicator of how much they feel safe. This concept was used in \cite{koay2005methodological, koay2006empirical, huttenrauch2006investigating,  kanda2008analysis, walters2008human, koay2009five, rodriguez2015bellboy, sorostinean2015reliable, rajamohan2019factors}.

An issue related to proxemics is that of human-robot approaching directions and spatial arrangements. An example of such a point of view, also drawn from psychology into pHRI, is that of \emph{F-formation}. During a conversation, people tend to place themselves approximately on a circle, and cooperate to maintain this specific spatial arrangement, for example by adjusting their spatial position and orientation as a new participant enters the conversation. These types of spatial arrangements, typical of conversations between several individuals, are referred to as \emph{F-formations}, as defined by Kendon \cite{kendon1990conducting}. The application of F-formations to pHRI and similar ideas on human-robot approaching directions and spatial arrangements were used in \cite{koay2005methodological, koay2006empirical, huttenrauch2006investigating, koay2006methodological, woods2006methodological, koay2007exploratory, koay2007living, walters2007robotic, koay2009five, neggers2018comfortable, rajamohan2019factors}.


\subsection{Direct input devices}\label{sec:devices}
The last category includes devices aimed at providing direct feedback during the experiment: these, as for questionnaires, provide a subjective measure of perceived safety, which however can be acquired in \lq\lq real-time'' rather than at the end of the experiment. The first device, used by Zoghbi et al. in \cite{zoghbi2009evaluation}, was named \emph{Affective-State Reporting Device} (ASRD), and was \lq\lq an in-house developed modified joystick [...] to record affective states expressed by each user'' \cite{zoghbi2009evaluation}. The second, used by Koay and coauthors in \cite{koay2005methodological, koay2006methodological, koay2006empirical, koay2007living, koay2009five} was the so-called \emph{Comfort Level Device} (CLD), i.e., \lq\lq a handheld comfort level monitoring device that would allow subjects to indicate their internal comfort level during the experiment'' \cite{koay2005methodological}.


\section{Industrial manipulators \cite{yamada1999proposal, hanajima2004influence, hanajima2005motion, kulic2005anxiety, kulic2006estimating, hanajima2006further, kulic2007affective, kulic2007physiological, zoghbi2009evaluation, arai2010assessment, aleotti2012comfortable, ng2012impact, lasota2014toward, weistroffer2014assessing, lasota2015toward, charalambous2016development, rahman2016trust, koppenborg2017effects, maurtua2017human, hocherl2017motion, you2018enhancing, pan2018evaluating, bergman2019close, koert2019learning, wang2019symbiotic, aeraiz2020robot, pollak2020stress, zhao2020task, hu2020interact}}\label{sec:IM}

\subsection{Overview on industrial manipulators}
Industrial manipulators constitute the type of robot on which the largest number of papers were written on perceived safety in pHRI. Earlier works used standard industrial manipulators, and precisely CRS A460 \cite{kulic2005anxiety, kulic2006estimating, kulic2007affective, kulic2007physiological, zoghbi2009evaluation}, Mitsubishi Movemaster RM-501 \cite{hanajima2006further,hanajima2004influence,hanajima2005motion}, ABB IRB-120 \cite{lasota2014toward, lasota2015toward}, Comau SMART SiX \cite{aleotti2012comfortable},  Yaskawa MOTOMAN-K10S and SONY SRX-410 \cite{ng2012impact}. More recent papers, instead, started using collaborative robots: KUKA LBR iiwa \cite{pan2018evaluating, maurtua2017human, pollak2020stress}, Kinova MICO \cite{rahman2016trust}, UR3 \cite{zhao2020task}, UR5 \cite{bergman2019close}, UR10, \cite{weistroffer2014assessing}, Sawyer \cite{hu2020interact}, and Franka Emika Panda \cite{aeraiz2020robot}. 

Although all considered works take perceived safety into account, they look at it from different points of view, as can be seen in the corresponding column of Table 1. 
Also, the concepts of valence and arousal were employed in \cite{kulic2006estimating, kulic2007affective, kulic2007physiological}, as part of a general framework used to describe the emotional experience of the participants. In addition to the previous categorization, papers \cite{yamada1999proposal, kulic2005anxiety, kulic2006estimating, kulic2007affective, kulic2007physiological, zoghbi2009evaluation, arai2010assessment, ng2012impact, lasota2014toward, lasota2015toward, charalambous2016development, rahman2016trust, maurtua2017human, hocherl2017motion, pan2018evaluating, bergman2019close, koert2019learning} focused on finding a connection between perceived safety, robot speed, and relative distance between human and robot, with \cite{aleotti2012comfortable, weistroffer2014assessing, koert2019learning} having an explicit focus on motion prediction. The human attitude toward robot position and approach direction was studied in \cite{hanajima2004influence, zoghbi2009evaluation}. Finally, the effect of visibility/audibility of the robot by the human subjects were analyzed in \cite{hanajima2004influence, hanajima2006further}.

\subsection{Selected works on industrial manipulators}
The focus of  Kulic and Croft \cite{kulic2005anxiety,kulic2007physiological} was the detection of anxiety and fear in the presence of a moving CRS 460 manipulator both via a questionnaire, and via physiological signals (heart rate, Galvanic skin response, corrugator muscle activity). Two motion planning algorithms were implemented, namely a potential field planner with obstacle avoidance, and a safe motion planner which added a danger criterion (i.e., the minimization of the potential force during a collision
along the path) to the above-mentioned potential field planner. As one can expect, the estimated human arousal increased when the robot was moving at higher speeds. Also, the subjects felt less surprised and anxious when the safe planner was used instead of the potential field planner, in particular when the robot was moving at high speeds.
Using fuzzy inference, an acceptable recognition rate was obtained with medium/high arousal levels, however this method proved itself unfit to provide a reliable estimate of valence. The reason for this is that the valence estimation rules relied on the corrugator muscle activity measurement, which (contrary to studies using picture viewing as stimulus for affective state response) became irrelevant when the robot motion was employed as stimulus.

As the method proposed in \cite{kulic2005anxiety, kulic2007physiological} was successful in estimating arousal, but unsuccessful in estimating valence, the next works of Kulic and Croft \cite{kulic2006estimating,kulic2007affective} introduced a user-specific method based on Hidden Markov Models (HMMs, see, e.g., \cite{ghahramani2001introduction}) for human affective state analysis. While running the same planners as in \cite{kulic2005anxiety, kulic2007physiological}, valence and
arousal were each represented by three HMMs, accounting for low, medium, and high level of valence/arousal. Contrary to the fuzzy inference engine proposed in \cite{kulic2005anxiety, kulic2007physiological}, user-specific HMMs could estimate valence using the collected aggregate physiological data. Also, the authors proved that attention played a big role in explaining measured physiological information because it showed whether the robot motion was the main stimulus of human emotions, or if they were caused by other environmental conditions.

Arai et al. \cite{arai2010assessment} aimed at measuring the subjects' mental strain while sharing their workspace with a moving industrial manipulator, both by using a SD questionnaire (evaluating fear, surprise, tiredness and discomfort) and by measuring Galvanic skin response. The subjects response was evaluated in relationship with relative distance with the robot, robot speed, and presence/absence of advance notification of robot motion. Based on the results of the experiments, it was concluded that, in order to avoid mental strain, a minimum distance of 2 m should be kept between human and robot, with the latter never exceeding a speed of 0.5 m/s. It was also observed that the presence of advance notification of the robot motions significantly contributed to reducing mental strain.

Charalambous et al. \cite{charalambous2016development} identified the main factors influencing trust in pHRI and generated a related trust measurement scale by running a study in two phases. They ran an exploratory study to gather participants' prior opinions so as to define a set of trust-related themes to generate a questionnaire, which was then used to obtain information from experiments conducted with three different industrial manipulators. It was concluded that the major determinants of safety-related trust development with industrial manipulators were the perceived threat given by the robot size, the absence of actual collisions with the operator, and the presence of a fluent and predictable robot motion (in particular, with the robot picking up objects at a low speed). More than half of the subjects noticed that having prior experience interacting with industrial robots would increase their confidence level.

Maurtua et al. \cite{maurtua2017human} conducted two experimental studies with a wide range of participants, gathering their responses on perceived safety via questionnaires, and concluded that both the safety systems that were implemented (a vision-based safety-rated monitored stop, and a power-and-force limiting strategy) were generally perceived as safe.

Koert et al. \cite{koert2019learning} focused on generating the robot motion via imitation learning with probabilistic movement primitives, by proposing two methods (based on spatial deformation and temporal scaling) for real-time human-aware motion adaptation. The main aim of the work was to guarantee perceived safety and comfort, using a goal-based intention prediction model learnt from human motions. Both methods were evaluated on a pick and place task with 25 non-expert human subjects, by analyzing motion data and using questionnaires on perceived safety and subjective comfort level. The predictability of the motion and the understanding of why the robot was responding in a certain way were always associated to higher levels of perceived safety: it was thus concluded that providing more communication (such as visual feedback) would probably enhance the perception of safety. However, the results showed that the subjects responses could hardly be generalized: for instance, the temporal scaling method was perceived as safe by one group of participants, and as unsafe by another group.

The work of Pollak et al. \cite{pollak2020stress} focused on stress during pHRI tasks with industrial manipulators. The experiments consisted of a collaborative task executed both in autonomous mode (in which the robot controlled all operations by itself) and in manual mode (in which each operation was initiated by the participant). The results on stress appraisal were collected both via questionnaires and via heart rate measurement. It was concluded that in manual mode the participants had higher levels of secondary stress appraisal (i.e., \lq\lq the complex evaluative process of what might and can be done about the demanding situation'' \cite{pollak2020stress}) and lower heart rates: this means that the participants  felt  less  confident (and thus more stressed) when not being in control (i.e. with the robot working in autonomous mode).

\subsection{Achieving perceived safety for industrial manipulators}
In general, for industrial manipulators, the feeling of perceived safety was enhanced when the relative human-robot distance was large \cite{arai2010assessment,weistroffer2014assessing,hocherl2017motion,you2018enhancing,bergman2019close}, possibly above a certain threshold (e.g., 2 m in \cite{arai2010assessment}), and when the robot speed was low \cite{kulic2005anxiety,kulic2007physiological,zoghbi2009evaluation,arai2010assessment,ng2012impact,zhao2020task}, also possibly below a certain threshold (e.g., 0.5 m/s in \cite{arai2010assessment}). A related result is that human subjects could accept relatively high acceleration and speed of the robot when their relative distance was larger \cite{yamada1999proposal,hanajima2005motion,lasota2015toward}. The robot size also influenced perceived safety, as smaller robots were perceived as less threatening \cite{charalambous2016development}.

Also, the robot motion should in general be fluent and predictable on its own \cite{charalambous2016development,hocherl2017motion,bergman2019close,koert2019learning,aeraiz2020robot,zhao2020task}, and/or should be made predictable by providing some type of communications to the human \cite{arai2010assessment,koert2019learning}. For this reason, the human subjects tended to be more comfortable if they could decide some aspects of the robot motion, i.e., when the motion will start \cite{pollak2020stress}: this is clearly related to the predictability issue highlighted in \cite{charalambous2016development,hocherl2017motion,bergman2019close,koert2019learning,aeraiz2020robot}. A related aspect was that, during a planned contact with the human (in the analyzed papers, due to the execution of an handover task), controlling forces and avoiding abrupt robots motions during handovers would also improve perceived safety \cite{aleotti2012comfortable,rahman2016trust,pan2018evaluating}.

The perception of safety improved when the participants had previous experience of the same task \cite{hanajima2006further,kulic2007affective,charalambous2016development,you2018enhancing,pan2018evaluating}, and/or when they had been previously informed about the robot safety features \cite{charalambous2016development,maurtua2017human}. According to the results reported in \cite{hu2020interact}, gender does not seem to play any role in the perception of safety, while extrovert participants tend to approach the robot at a closer distance.

\section{Mobile robots \cite{koay2005methodological, koay2006empirical, huttenrauch2006investigating, koay2006methodological, woods2006methodological, syrdal2006doing, dautenhahn2006may, walters2007robotic, walters2008human, syrdal2009negative, karreman2014robot, bhavnani2020attitudes, scheunemann2020warmth}}
\subsection{Overview of works on mobile robots}
As mobile robots, we consider a moving base without robot arms or other parts moving on it, so that the safety of the motion only relates to the base itself. The most popular robot in the analyzed papers was PeopleBot (ActivMedia Robotics) \cite{koay2005methodological, koay2006empirical, koay2006methodological, huttenrauch2006investigating, woods2006methodological, syrdal2006doing, dautenhahn2006may, walters2007robotic, walters2008human, syrdal2009negative}, followed by Giraff \cite{karreman2014robot}, Cozmo \cite{bhavnani2020attitudes} and Sphero \cite{scheunemann2020warmth}. 
The aim of most of the described studies is to establish comfort level as the robot approaches the participants or moves around them, without explicit focus on the perceived possibility of injuries that was instead present when dealing with industrial manipulators. This could be explained by the fact that the considered mobile robots move at relatively low speeds, and thus are not perceived as a threat in terms of possible physical harm.

Mobile robots were employed for service, transferring, entertainment and navigation tasks, such as fetch and carry \cite{koay2005methodological, koay2006empirical, woods2006methodological, walters2007robotic, syrdal2009negative, dautenhahn2006may}, following humans \cite{huttenrauch2006investigating, koay2006methodological}, being reached by or reaching humans \cite{walters2008human, karreman2014robot, dautenhahn2006may, bhavnani2020attitudes}. When the participants chose the most and least preferred approach directions of the robot motion \cite{koay2005methodological, koay2006empirical, huttenrauch2006investigating, woods2006methodological, syrdal2006doing, dautenhahn2006may, walters2007robotic, syrdal2009negative, karreman2014robot}, typical concepts related to proxemics \cite{huttenrauch2006investigating, walters2008human, scheunemann2020warmth, bhavnani2020attitudes} and F-formation \cite{huttenrauch2006investigating, karreman2014robot} were employed, in particular analyzing how close to the human the robot could get, and from what direction. Robot speed and distance were directly considered in \cite{woods2006methodological, dautenhahn2006may, syrdal2006doing, walters2007robotic, walters2008human, syrdal2009negative, scheunemann2020warmth}.


\subsection{Selected works on mobile robots}
Koay et al. \cite{koay2005methodological} studied the correlation between participants comfort and their distance with the robot, while the latter was moving around them. The participants feedback was collected via the CLD already introduced in Section~\ref{sec:devices}. The subjects were typically showing discomfort when the robot would either move behind them, or block their path, or be on a collision route with them. Similar conclusions were obtained by Koay et al. \cite{koay2006empirical}, in which the experiments were recorded on video, later reviewed to assess human comfort based on body movements, body language, facial expressions, and speech. 


During the experiments described by H\"{u}ttenrauch et al. \cite{huttenrauch2006investigating}, the human subjects conducted the tour of a room with the robot, while the latter moved after them; then, the human asked the robot to search for specific pieces of furniture or to close/open them. The information from the subject's voice would be heard by a human operator, who would remotely guide the robot with a \emph{Wizard-of-Oz} approach. After the experiments, the researchers conducted a spatial interaction analysis via users' questionnaires, videos, and voice recordings of the users' commands. In particular, the analysis of the subjects' reactions to the relative distance with the robot was based on Hall’s proxemics, while the relative positioning of human and robot (e.g., robot behind or in front of the subject) was analyzed in terms of F-formation. The conclusions of the study showed that the participants preferred a distance to the robot in the $0.45-1.2$ m range (i.e., in the \emph{personal} zone in terms of Hall’s proxemics), and a relative face-to-face positioning in terms of F-formation arrangement. 


Woods et al. \cite{woods2006methodological} took into account different initial positions of the subjects, while the robot was handing them a snack. The users provided their preferences on the robot motion through a questionnaire after each experimental trial. In addition to the subject who would directly interact with the robot (\emph{live trial}), another subject was present, watching a live video streaming of the scene (\emph{video trial}): this second participant also had to provide feedback on the robot motion. In the first scenario (subject sitting at a table), live subjects preferred the robot approaching them from the front-left or front-right direction, whereas the subjects in video trials gave preference to the robot approaching from the front. In the second scenario (humans standing against a wall positioned behind them), subjects in both live and video trials felt uncomfortable when the robot was moving towards the participant from the front. Finally, in scenarios with the subject sitting on a chair or standing (in both cases, in an open space), both subjects in live and video trials preferred the robot approaching from the front-left and front-right directions (as in the first scenario), while the robot approaching from the back was the least comfortable case.

By considering a similar experimental setup and scenario, the work by Syrdal et al. \cite{syrdal2006doing} aimed at finding a correlation between the subjects' comfort level as PeopleBot was approaching them from different directions, and their personality traits. These were defined with reference to the Big Five Domain Scale described in Section \ref{sec:assess}. The experiment results showed that the most comfortable directions of robot motion for the human subjects (evaluated on a Likert scale) were front-right and front-left. Interestingly, the subjects' personality traits did not affect, on average, their preferences of robot approach direction; however, more extrovert subjects showed higher rates of tolerance to robot behaviour when the robot approached them from \lq\lq uncomfortable'' directions, such as from behind.

The work by Dautenhahn et al. \cite{dautenhahn2006may} describes the results of two studies which
investigated the best approach direction to a seated human participant. Two experimental trials were run in which the robot fetched an object requested by the human, coming from different approach directions. The majority of subjects felt more comfortable with the robot approaching from the right or the left side, rather than from the front.

In the work of Walters et al. \cite{walters2008human}, experiments were conducted in which the subjects would interact with PeopleBot. Based on Hall's proxemics theory, it was observed that $56\%$ of the participants let the robot enter their personal zone. Also, the human subjects who had previous experience of working with PeopleBot approached it, on average, at a closer distance ($51$ cm) than those with no previous experience ($73$ cm). In both cases, these distances were in the range typically used for human-human interaction between friends and family members, i.e. $40$-$80$ cm (see, e.g., \cite{nakauchi2002social}). 

The experiments described by Syrdal et al. \cite{syrdal2009negative} consisted of two interaction sessions with PeopleBot. In one session the robot presented a more \lq\lq socially interactive'' behavior than in the other (i.e., the robot adapted its behavior to the participants, rather than treating them as any other obstacle in the environment). The conclusions showed that the more \lq\lq socially interactive'' behavior of the robot was not comfortable for the user, due to the lower level of predictability of the robot motion.


\subsection{Achieving perceived safety for mobile robots}
When the mobile robot approached the subjects, the front-right and front-left directions of approach seemed to be the most comfortable, and the approach from behind the most uncomfortable \cite{koay2005methodological,woods2006methodological,syrdal2006doing,karreman2014robot}, with participants with a higher degree of extroversion showing a higher tolerance to uncomfortable directions of approach \cite{syrdal2006doing}. The direct frontal approach was not perceived as comfortable when the subjects were sitting on a chair or leaning against a wall, while it was considered acceptable in open spaces \cite{woods2006methodological,dautenhahn2006may,walters2007robotic}. When the robot was following them, the subjects preferred it to be at the side rather than perfectly behind them, probably in order to maintain it within their field of view \cite{koay2006methodological,walters2007robotic}. The participants showed low levels of comfort when the robot was either blocking their path, or was on a collision route with them \cite{koay2005methodological,koay2006empirical}. 

When executing a task in cooperation, the preferred distance was in the personal zone in terms of Hall's proxemics, for which we give an estimate of $46-80$ cm from the human based on the intersection of the results in \cite{huttenrauch2006investigating,walters2008human}. However, when subjects were executing an independent task, they could feel discomfort if the robot was closer than 3 m, i.e., in the social zone reserved for human-human face to face
conversation \cite{koay2006empirical}. Participants from certain cultural backgrounds were also found to feel comfortable with the robot at a closer distance \cite{bhavnani2020attitudes}. In terms of F-formation, the face-to-face arrangement was shown to be the preferred one when executing a task in cooperation \cite{huttenrauch2006investigating}.

Subjects with previous pHRI experience allowed the robot to come closer, as compared to subjects without previous experience \cite{walters2008human}. Finally, the human subjects were uncomfortable when the robot motion was unpredictable \cite{syrdal2009negative}, or when the robot had a large size \cite{bhavnani2020attitudes}.

\section{Mobile manipulators \cite{inoue2005comparison, dehais2011physiological, chen2011touched, strabala2013toward, dragan2015effects, brandl2016human, macarthur2017human}}
\subsection{Overview of works on mobile manipulators}
In this survey paper, we consider a robot as a mobile manipulators when it consists of one or more robotic arms mounted on a moving base. We exclude from this category the papers in which the robot has a face (even if realized, e.g., by using a computer screen), and we include them in the category of humanoid robots, described in the following section. The following mobile manipulators were used in the articles: Jido \cite{dehais2011physiological}, HERB \cite{strabala2013toward}, and Care-O-bot 3 \cite{brandl2016human}.

Similarly to industrial manipulators, the effect of the speed of mobile manipulators and of their distance with respect to human subjects was analyzed in several articles \cite{dehais2011physiological, strabala2013toward, brandl2016human, macarthur2017human}. The experiments described in \cite{inoue2005comparison} were conducted using virtual reality tools, while the other described articles made use of real-world robots.


\subsection{Selected works on mobile manipulators}
Dehais et al. \cite{dehais2011physiological} used a previously-developed human-aware motion planning algorithm to provide the robot with safe and ergonomic movements while executing a bottle handover task with a human companion. Three different types of motion were executed, which varied in terms of use of the planner, grasp detection, and speed. The levels of stress and comfort (defined, in this specific case, as the physical demand required to grasp the bottle) were evaluated both via questionnaires and by monitoring Galvanic skin response, deltoid muscle activity and eye gaze. Focusing on the results regarding the robot speed, the robot motion with medium velocity was judged as the safest and most comfortable. Motions with high speed were perceived as the most unsafe, while motions with low speed still had low levels of comfort, as the participants would try to grasp the bottle before the robot motion was concluded.

The aim of Chen et al. \cite{chen2011touched} was to understand the human reaction to robot-initiated touch. In order to do that, experiments were conducted with 56 subjects, in which the Cody robot would touch and wipe their forearms. The robot could either verbally warn the subject before contact or not. Also, it could verbally state if the touch had the aim of cleaning the participant's skin, (instrumental touch) or to give comfort (affective touch), even though the robot motion would be exactly the same. As a first main result, it was found that participants felt more comfortable when believing that  an instrumental touch, rather than an affective touch, was being performed. This demonstrated that ``the perceived intent of the robot can significantly influence a person’s subjective response to robot-initiated touch'' \cite{chen2011touched}. The second main result was that participants showed a higher level of comfort when no verbal warning was present. Even if this did not lead to any clear conclusion, it showed that verbal warnings are not necessarily improving the participants experience, and should be carefully designed.

Strabala et al. \cite{strabala2013toward} first reviewed studies on how an handover task is executed between humans, with focus on their
coordination process, in particular in terms of exchanged signals and cues. Based on these studies, they established a framework that considered the way in which humans \lq\lq approach, reach out their hands, and transfer objects while simultaneously coordinating
the \emph{what, when, and where} of handovers: to agree that the handover will happen (and with what object), to establish the timing of the handover, and to decide the configuration at which the handover will occur.'' \cite{strabala2013toward}. They then proposed a coordination framework for human–robot handovers that separately took into account the physical and social-cognitive aspects of the interaction, and finally evaluated human–robot handover behaviors via experiments. It was concluded that the human participants did not feel safe and comfortable when the robot applied a high force when handling an object to them, or when it maintained a high speed when close to their hands, or when it was executing a motion that was in general perceived as unpredictable.

The focus of the paper of Dragan et al. \cite{dragan2015effects} was on different characteristics of robot motion and how they influence physical collaboration with human subjects. A robot motion can be only functional (i.e., it  achieves the target without collisions), or, in addition, predictable (i.e., it meets the collaborator’s expectancy given a known goal), or legible (i.e., it enables the collaborator to confidently infer the goal). Experiments were conducted in which a mobile manipulator collaborated with a human participant to prepare a cup of tea, and an important aspect was how the human perceived the motion of the robot while the latter was moving to grasp the cup. All three types of motion (functional, predictable, and legible) were executed, and a questionnaire was filled after each trial. In conclusion, both predictable and legible motions were perceived as safer than functional-only motions. This was due to the fact that a legible motion was better at conveying intent, but this did not lead to an increased level of perceived safety and comfort as compared to a motion that was only predictable.

\subsection{Achieving perceived safety for mobile manipulators}
Perceived safety improves with human-robot distance \cite{macarthur2017human}, and if the robot motion does not happen at a high speed, especially when close to the human body, in particular to face or hands \cite{dehais2011physiological,strabala2013toward,macarthur2017human}. According to the findings in \cite{brandl2016human}, the participants allowed the robot to approach them at a closer distance (specifically, $57$ cm) if the robot was moving slowly, and/or if they had previous experience of the same task. Additionally, the predictability of the motion increased the level of perceived safety \cite{dehais2011physiological,strabala2013toward,dragan2015effects}. Specifically for handover tasks, the feeling of safety would decrease if the robot applied a large force on the object, or transferred it slowly and did not release it for a relatively long time \cite{dehais2011physiological,strabala2013toward}. 

\section{Humanoid robots \cite{butler2001psychological, kanda2002development, sakata2004psychological, nomura2004psychology, itoh2006development, koay2007exploratory, edsinger2007human, koay2007living, huber2008human, kanda2008analysis, huber2009evaluation, koay2009five,  takayama2009influences, bainbridge2011robot, de2013relation, chan2013human, moon2014meet, rodriguez2015bellboy, sorostinean2015reliable, haring2015changes, sadrfaridpour2017collaborative, sarkar2017effects, munzer2017impact, ivaldi2017towards, willemse2017affective, neggers2018comfortable, charrier2018empathy, stark2018personal, rajamohan2019factors, block2019softness, fitter2019does, busch2019evaluation, fitter2020exercising, dufour2020visual}}

\subsection{Overview of works on humanoid robots}
The articles presented in this section focus on humanoid robots, i.e., robots with human-like upper body part, together with either legs or a base, that in turn can be either mobile or stationary. As such, the following robot platforms were included in this section: Robovie \cite{kanda2002development, nomura2004psychology, kanda2008analysis}, HRP 2 \cite{sakata2004psychological, bainbridge2011robot}, WE-4RII	\cite{itoh2006development}, Nao \cite{de2013relation, willemse2017affective}, PeopleBot with head \cite{koay2007exploratory}, Domo \cite{edsinger2007human}, iCat with arms \cite{huber2008human}, JAST \cite{huber2009evaluation}, Willow Garage PR2 \cite{takayama2009influences, chan2013human, moon2014meet, block2019softness}, Meka \cite{sorostinean2015reliable}, Robi \cite{haring2015changes}, Baxter \cite{stark2018personal, sadrfaridpour2017collaborative, munzer2017impact, sarkar2017effects, fitter2019does, fitter2020exercising, dufour2020visual}, iCub \cite{ivaldi2017towards}, Pepper \cite{neggers2018comfortable, charrier2018empathy}, and ARMAR-6 \cite{busch2019evaluation}. Additionally, there were also works that focused on comparison of two or more robot platforms with each other: PeopleBot with four appearance configurations \cite{koay2007living, koay2009five}, Robovie with ASIMO \cite{kanda2008analysis}, Sacarino with or without a human-like upper part \cite{rodriguez2015bellboy}, Nao with PR2 of two height configurations \cite{rajamohan2019factors}, Nomadic Scout II with and without a mock-up body \cite{butler2001psychological}. 

In addition to perceived safety, works exploiting humanoid robots often investigate the perception of humanlikeness or anthropomorphism of the humanoid robots \cite{bainbridge2011robot, sakata2004psychological, kanda2008analysis, chan2013human, haring2015changes, willemse2017affective, rajamohan2019factors, busch2019evaluation}, compare robot and human conditions in their experiments \cite{kanda2008analysis, huber2008human, huber2009evaluation}, and are inspired by psychological findings of human-human interaction \cite{takayama2009influences,chan2013human, moon2014meet, ivaldi2017towards, charrier2018empathy, rajamohan2019factors, block2019softness}.

The tasks that the robot executed in close distances with the human subjects were: pick and place \cite{sakata2004psychological}, handover  \cite{edsinger2007human, huber2008human, stark2018personal, chan2013human, moon2014meet, huber2009evaluation}, hand-shaking \cite{bainbridge2011robot}, talk and touch \cite{de2013relation, willemse2017affective}, approaching the human subject \cite{butler2001psychological, takayama2009influences, neggers2018comfortable, rajamohan2019factors, koay2007living, koay2009five}, or playing games \cite{bainbridge2011robot, charrier2018empathy, koay2007living, koay2009five}.

\subsection{Selected works on humanoid robots}
Robovie was used in several earliest works investigating psychological safety  \cite{kanda2002development, nomura2004psychology, kanda2008analysis}. Kanda et al. \cite{kanda2002development} presented the Robovie robot's development details and one of its first evaluations with people. The experiment compared three robot responses to human-initiated interactions: passive (after touching, the robot performed one friendly action and went into standby mode), active (after touching, the robot displayed active friendly behavior, and the participant observed it), and complex (after touching, the robot went to \emph{idling} and \emph{daily work}, i.e., move around, behaviors). The participants preferred  passive behavior and stayed away from the robot at a distance of 41 centimeters (i.e. intimate zone). 

Another robot that was also exploited for similar research questions was PeopleBot  \cite{koay2007exploratory, koay2007living, koay2009five}.  Koay et al. in \cite{koay2007living} and \cite{koay2009five} presented the results of a long-term experiment (5 weeks, 8 sessions) with 12 participants. Each group of participants (two male and one female) interacted with one of four variations of PeopleBot: small/tall mobile robots with no head or small/tall humanoids. There were three different approaching scenarios (physical, verbal and no-interaction). The comfort level of the participants was evaluated both via CLD (similarly to \cite{koay2005methodological, koay2006methodological, koay2006empirical}), a questionnaire and a semi-structured interview based on the Big Five domain scale \cite{de2000big}. It was concluded that the robot approaching from the front and side was evaluated as the most appropriate. Also, the participants were more likely to allow the robot to come closer at the end of the 5-week period rather than at the beginning. In addition, as time passed, participants preferred verbal interaction to physical interaction, which was not the case at the beginning of the experiment. In \cite{koay2009five}, it was concluded that humans were not feeling comfortable when the robot blocked their path, or moved on a collision route towards them (either from the front or from behind). The participants felt comfortable when the robot notified them before moving closer. It was also concluded that the predictability of the robot motion would greatly increase the level of participants' comfort. 

The handover task performed by the iCat robot embedded with arms was evaluated in an experiment by Huber et al. \cite{huber2008human}, which tested two velocity profiles for the robot: a conventional trapezoidal velocity profile in joint coordinates and a minimum-jerk profile of the end-effector in Cartesian coordinates. Participants rated how human-like the robot motion was and how safe they perceived it. The results suggest that participants' level of trust was similar in both cases, and both velocity profiles were similarly perceived as human-like. However, the minimum-jerk profile had higher safety ratings, and the robot was perceived as safe if its maximum speed did not exceed 1 m/s. 

Additionally, Willow Garage PR2 was also used to investigate perceived safety \cite{takayama2009influences, chan2013human, moon2014meet, block2019softness}. Takayama et al. \cite{takayama2009influences} investigated various factors that influence proxemic behaviors and perceived safety with PR2. The experiment consisted of three conditions: the human approaching the robot, and an autonomously moving or teleoperated robot approaching the human. Findings from observing people's behaviors and their answers in the questionnaire suggest that those people that own pets or have experience with robots let the robot come closer. Additionally, when the robot faced downwards looking at the participants' legs, participants of both genders came close to the robot, whereas women stood further away in comparison to men in cases when the robot looked at their faces. Finally, the results based on NARS suggest that people who had negative attitudes towards robots felt less safe and less comfortable being closer to them. Inspired by the goal to give PR2 an ability to perform an object handover mission with non-expert users in unpredictable surroundings, Chan et al. \cite{chan2013human} designed four different robot controller configurations obtained by tuning the initial grip force and the release force threshold. During the experiments the robot delivered an instrumented baton to a human subject using the following four control configurations: human-like, balanced, constant-grip-force and quick release. The questionnaire given to humans after each trial solicited questions on PR2's motion safety, efficiency and intuitiveness. The overall results showed that the human-like configuration was the preferred one, while the constant-grip-force controller received the worst evaluation because it was less predictable for the human subjects. Moon et al. \cite{moon2014meet} designed and evaluated attention gaze cues of the PR2 robot during a handover with the aim to improve the user experience during task execution. Three gaze patterns were evaluated by the participants: no gaze (robot's constant head position), shared attention gaze (robot's staring at the transferring object) and turn-taking gaze (looking at the subject). All participants were assigned to two random gaze conditions and compared them by answering questions on overall preferences, naturalness and timing communication. In case of a shared attention gaze, the human subjects reached for the object earlier than in the other two cases, while the results suggest that participants preferred handovers to have a turn-taking gaze approach in comparison to other methods. 

Butler and Agah \cite{butler2001psychological} focused on the psychological impact of robot behavior patterns on humans in daily life, such as approaching or avoiding a human and executing tasks in crowded surroundings. Two types of robot were used in the experiments: mobile robot and humanoid. Participants rated the fast moving humanoid with maximum speed of 40 inches/sec as the most uncomfortable and a motion with slow speed (10 inches/sec) for both robot types as the most pleasant. Overall, when the body was added to the mobile base and the robot became a humanoid, then the human subjects rated the behavior of such robot as less comfortable.  

A comparison of humanoid robots' appearances was performed by Kanda et al. \cite{kanda2008analysis}, where Robovie with a wheel mechanism and ASIMO with a biped-walking mechanism were compared to a human in an experiment consisting of a greeting, a small talk, and a guiding behavior. Politeness and proxemics maintained by the participants were acquired via behavioral assessment and SD questionnaire. The results suggest that people were similarly polite in all conditions, while they kept the closest distance from ASIMO in comparison to another human or Robovie.

Rajamohan et al. \cite{rajamohan2019factors} studied the influence of various factors on people's preferred distance during human-robot interactions. Participants were asked to say ``stop'' when they felt uncomfortable when one of the robots (a humanoid Nao robot (58 cm), a short PR2 (133 cm), or a tall PR2 (164.5 cm) approached them. They were also asked to approach the robots themselves and stop when they felt uncomfortable. The results suggest that people prefer to be approached by the robot while men allowed robots to come closer than women did. The height of the robots had also effected the maintained distance, as the Nao robot was allowed to come closer to people in comparison to both PR2 robots. In addition, it was selected as the most comfortable to interact with. 

Perceived safety and trust were the main criteria in the work of Fitter et al. \cite{fitter2019does}, where a human and a Baxter robot played a hand-clapping game for an hour to evaluate robot's facial reactivity, physical reactivity, arm stiffness, and clapping tempo. Baxter was perceived as more pleasant and energetic when facial reactivity was present, while physical reactivity made it less pleasant, energetic, and dominant for participants. Higher arm stiffness increased perceived safety and decreased dominance while faster tempo of clapping made Baxter seem more energetic and more dominant. 

\subsection{Achieving perceived safety for humanoid robots}
The majority of works involving humanoid robots investigated the effect of people's proxemics with the robot and their preferred approaching direction. As such, there are findings suggesting that participants preferred Robovie to stay as close as 41 centimeters from them \cite{kanda2002development}. People had a strong preference for being approached from the front by PeopleBot \cite{koay2009five}, also demonstrating that a developed cohabituation effect influenced these preferences. Along these findings, a prior experience with pets and robots also demonstrated an effect on people's proxemics with PR2 \cite{takayama2009influences} and PeopleBot \cite{koay2007living}.  

Perceived safety was also explored comparing two or more robots. As such, the small Nao was allowed to come closer by participants in comparison to both short and tall PR2s \cite{rajamohan2019factors}; people kept the shortest distance from ASIMO in comparison to Robovie or another human \cite{kanda2008analysis}; PeopleBot's mechanoid or humanoid appearances had stronger effect than the robot height \cite{koay2007living}. The robot motion was also a focus in other works in this section: Baxter's higher arm stiffness increased perceived  safety \cite{fitter2019does}, the level of comfort increased when the robot motion was more predictable \cite{koay2009five}, and a minimum-jerk profile of the iCat arms had higher safety ratings \cite{huber2008human}.

Additionally, non-verbal social cues were also explored: PR2's eye contact was important for handover tasks \cite{moon2014meet}, while Baxter's facial reactivity was perceived as more pleasant \cite{rajamohan2019factors}. Gender effects have also been found with men letting the robot come closer than women did \cite{rajamohan2019factors}, in particular in situations when the robot looked directly into the participants' face \cite{takayama2009influences}.

\section{Drones \cite{duncan2013comfortable, szafir2014communication, cauchard2015drone, jones2016elevating, duncan2017effects, yeh2017exploring, acharya2017investigation, karjalainen2017social, chang2017spiders, abtahi2017drone, jane2017drone, colley2017investigating, kong2018effects, jensen2018knowing, yao2019autonomous, wojciechowska2019collocated}}

\subsection{Overview of works on drones}
Flying robots have increasingly been used in the last years for various applications that involve humans (the first considered paper, i.e., \cite{duncan2013comfortable} only dates back to 2013). Most papers that analyzed this robot type made use of drones (often referred to as small unmanned aerial vehicles, or sUAV in short) for several tasks, the most common being aerial photography, delivery, monitoring, and mapping. 

The following UAVs were utilized by the studies discussed in this section: AIR-Robot-100B \cite{duncan2013comfortable}, DJI Phantom 2 \cite{cauchard2015drone, szafir2014communication} and 3 \cite{yeh2017exploring, colley2017investigating}, Parrot Bebop \cite{jones2016elevating} and 2 \cite{jensen2018knowing}, Parrot AR.Drone \cite{szafir2014communication, chang2017spiders, abtahi2017drone} and 2 \cite{wojciechowska2019collocated}, AscTec Hummingbird \cite{acharya2017investigation}, and Georgia Tech Miniature Autonomous Blimp \cite{yao2019autonomous}. There were also works evaluating simulated drones with the help of the HTC Vive VR headset \cite{karjalainen2017social} and the Cave Automatic Virtual Environment (CAVE) setup \cite{duncan2017effects}. 


\subsection{Selected works on drones}
One of the earliest studies with drones was conducted in 2013 by Duncan et al. \cite{duncan2013comfortable}, who investigated how close would participants allow a drone to approach them before feeling uncomfortable or anxious, and whether that distance would change based on the height (altitude) of the drone. To this end, AIR-Robot-100B was mounted on a platform fixed at the ceiling that controlled the approaching height (2.13 m vs 1.52 m), angle, and speed. Participants were standing at the taped location facing the robot that would slowly move to them. A within-subject study with 16 participants did not produce statistically significant findings suggesting that there were no effects measured with either a stop distance metric or a heart rate variability physiological metric (BIOPAC). The authors suggest that a potential confounding factor might have been the need to tell the participants that they are safe per ethical board requirements. 

Szafir et al. \cite{szafir2014communication} designed four high-level signal communication mechanisms (blinker, beacon, thruster, and gaze) embedded onto two drones (Parrot AR.Drone 2.0 and DJI Phantom 2 drones) in order to convey drone flight intentions. They were evaluated by 16 participants in the lab study that confirmed their efficacy for intent communication, which in turn increased perceived safety. 

One of the first outdoor studies with the DJI Phantom 2 drone and 19 volunteers was conducted by Cauchard et al. in 2015 \cite{cauchard2015drone}. It was a user-defined interaction elicitation study where participants had to instruct the drone to perform 18 tasks using voice or gestures (e.g. fly closer, follow, take a selfie, etc.), which were then executed in a Wizard-of-Oz fashion. People generally treated the drone as if it were an animate being: a person, a group of people, or even a pet. Most participants reported feeling safe and were more concerned with the drone's safety. Most participants were comfortable with bringing the robot into their intimate (N=7) and personal space (N=9). Propeller noise and generated wind created discomfort with several participants wishing for an emergency landing interface. 

Jane et al. \cite{jane2017drone} replicated the Cauchard et al. \cite{cauchard2015drone} study in China with 16 participants and compared their findings to understand if there were any cultural differences. Most participants in China (50\%) allowed the drone to be within their intimate zone, while the majority of US participants (47\%) were more comfortable with personal space for their  interaction with the drone. Additionally, participants in China were significantly less likely to compare the drone to a pet than those in the US. 

Jones et al. \cite{jones2016elevating} studied the use of a semi-autonomous Parrot Bebop drone for shared video-conferencing tasks between an outdoor user that was collocated with a drone and a remote user who could explore the environment from the drone's perspective. After eight pairs of participants collaborated on shared navigation and search tasks, people's responses in the interviews suggested that collocated field participants were generally more concerned about the drone and its safety than for their own safety. Additionally, participants concerns were related to balancing the size of the drone and its proximity to collocated people. While most of them reported feeling comfortable around the drone, some field participants reported feeling somewhat disturbed by the noise. In their conclusions, the authors suggested to communicate safety messages related to drone's malfunctioning, taking off and landing, battery life, and close proximity to obstacles.

To give people a chance to form a realistic mental model of a drone, Chang et al. \cite{chang2017spiders} conducted a study with 20 participants that interacted with either a real Parrot AR.Drone or a life-sized black cardboard prototype of a drone to clearly isolate the features of a real drone (such as sound, wind, and speed) that typically affect privacy and security concerns. Apart from being of the same size, shape, and color as the real drone, the cardboard drone also mimicked the real drone's movements and had features resembling a camera. It was found that sudden speeding up, unstable maneuvers and unusual movements such as flips and back-and-forth movements were perceived as threatening and  potentially dangerous. Although the wind and sound produced by the real drone were viewed as negative, people also commented they made the drone appear less stealthy. Participants also shared concerns with drones being near people, buildings, other drones, and wildlife. Finally, it was suggested to communicate the purpose of a drone via its design using colors, logos and decorations, to balance its size and shape (with a circular shape being less threatening), and to stabilize and engineer drone movements to make them appear as safe and user-friendly. 

In a study by Duncan and Murphy \cite{duncan2017effects}, 64 participants navigated around the CAVE as the first person avatar and interacted with a simulated drone that performed behaviors inspired by birds and other animals (e.g. silently circling overhead). Results suggest that the low speed of 2 m/s as well as high cyclicity (i.e. repetition) expressions can be used to increase distancing in human-drone interactions. 

Informed by the human-centered interaction design that involved a design study and a focus group, Yeh et al. \cite{yeh2017exploring} built a social drone by adding a blue oval-shaped safety guard, an android tablet displaying a face, and a greeting voice to the DJI Phantom 3 drone. Their findings suggest that a social drone is significantly better at closing the distance in comparison to a non-social drone (a square-shaped drone and no face.) i.e. a social drone was allowed to come as close as 1.06 meters on average which is 30\% closer than a non-social drone with no voice. Female participants stopped the drone at a distance of 1.5 meters on average, while male participants let the drone fly as close as 1.1 meters. Additionally, people were more comfortable with the height of 1.2 m as their average distance was 1.14 m in comparison to 1.35 m for the height of 1.8 m. A bigger lateral distance of 0.6 m helped decrease the distance for the drone. The results also demonstrated that drone noise added to the mental stress of the participants while they still preferred medium-sized drones for visibility. 

Karjalainen et al. \cite{karjalainen2017social} designed and evaluated a companion drone for a home environment in a three-staged user-centered approach. As the final stage of their design process, they conducted a study with 16 participants in a room adapted for virtual reality with the help of the HTC Vive VR headset. Their results suggest that the preferred drone appearance was a round shape that has a face. 

Acharya et al. \cite{acharya2017investigation} compared an AscTec Hummingbird drone to a mobile Double Telepresence robot and measured the distances at which 16 participants stopped the robots. There were significant differences found: the mean distance was 65.5 cm from the Hummingbird and 36.5 cm from the Double.

Abtahi et al. \cite{abtahi2017drone} built a safe-to-touch drone by adding a light-weight wooden (balsa, bamboo, and basswood) frame and a clear polypropylene mesh to prevent direct contact with the blades of the Parrot AR.Drone. In their between-subjects study with 24 participants, all participants that interacted with the safe-to-touch drone did it in their intimate space ($<$ 0.45 m) compared to 42\% in the control conditions (i.e. unsafe drone without the safety guards around propellers). In addition, the minimum distance between the participants and the safe-to-touch drone was significantly  less than the minimum distance in the control condition. 58\% of participants touched the safe-to-touch drone and 39\% of all interactions were touch-based. The participants also reported that safe-to-touch drones were perceived as significantly less mentally demanding, while 83\% of participants reported feeling safe when interacting with this drone in comparison to only 42\% in the control condition, though the difference was not significant. Additionally, it was noted that people tended to animate drones and compared them to pets.

Jensen et al. \cite{jensen2018knowing} designed and evaluated four drone gestures (waggle, nod, toss, and orienting) in order to signal acknowledgement to an interacting human partner. The Parrot Bebop 2 drone was used in a study with 16 participants who were approached by the drone and approached the drone twice for a total of four encounters. The starting proxemics was 9.1 m apart at the altitude of 1.5 m and a speed of 0.7 m/s. People were asked to indicate the desired acknowledging distances. The result was 1.8 m on average for all participants with no significant differences between conditions. The participants found the drone's flight altitude and the speed at which it approached them about right. People approached the stationary drones on average at a slightly higher speed (0.9 m/s). In their second study, four gestures were evaluated with the orienting gesture making people feel significantly more acknowledged than nod, toss or waggle. Rotating the drone towards the user elicited a higher degree of acknowledgement. Gestures were effective at communicating their intent at the distance of four meters. Some of the participants experienced some of the changes in drone speeds as potentially threatening, aggressive, or erratic behaviour. 

Kong et al. \cite{kong2018effects} conducted a study with 60 participants that were given a 4-minute tour by either a human-driven or by an algorithm-driven drone around a university campus. Participants were provided with earphones with an audio guide that had an explicit statement that the drone was either controlled by a human or fully autonomous, but the study was always controlled in a Wizard-of-Oz fashion. Additionally, the drone was also manipulated in terms of safety with unstable and stable flying behaviors. Their findings suggest that there were no differences between robot control conditions, while perceived safety  significantly affected participants' satisfaction. Additionally, there was an interaction effect between type of drone control and level of perceived safety: when the drone was safe, people were highly satisfied regardless of who controlled the drone, whereas people were significantly dissatisfied when the drone was human-driven but flying unsafe. In a similar walking tour study, Colley et al.  \cite{colley2017investigating} was guided by a flying drone that communicated the route using its movements. The participants' (N=10) preferred distance from the drone was 4.0 m, and its flying height was 2.6 m. 

The study by Wojciechowska et al. \cite{wojciechowska2019collocated} explicitly asked 24 participants to sort their preferred interaction with the Parrot AR.Drone 2.0 after they experienced 12 approaching trials: three proximity cases (intimate, personal and social), three trajectories (straight, up-to-down and down-to-up), three speed parameters (slow, moderate and fast) and three directions (front, front-side and rear) of drones. The results of this study revealed that people had significant preferences for the personal zone (1.2 m), for a moderate speed (0.5 m/s on average), for the frontal side of approach, and for a straight trajectory. In addition, there was an interesting observation suggesting that the invasion of an intimate zone by the drone results in decline in people's comfort level, which remained low even after the proximity of the interaction increased again.  Additionally, participants felt safe despite the fact that they were aware of the lack of autonomous drone control, and participants often identified drones with pets and attributed human behavior to them \cite{wojciechowska2019collocated}. 

In order to explore human-blimp interaction, Yao et al. \cite{yao2019autonomous} studied how comfortable the LED display feedback from the blimp to the user was. To this end, they designed and built the Georgia Tech Miniature Autonomous Blimp (GT-MAB) that was evaluated in a study with 14 participants. Their results showed that the majority of the participants were able to successfully lead the autonomous blimp using gestures. Most participants agreed that the LED visual feedback provided a better interactive experience as it seemed like the blimp was ``thinking.''

\subsection{Achieving perceived safety for drones}
An increasing body of research is examining issues of perceived safety within physical human-drone interaction. Those works that were concerned with drones' proximity suggest that people feel safe when drones are positioned within personal zone at a distance of 65.5 cm \cite{acharya2017investigation}, 1.06 m \cite{yeh2017exploring}, and 1.2 meters \cite{cauchard2015drone, wojciechowska2019collocated} and that people prefer to be acknowledged by drones at a distance of a minimum of 1.8 meters \cite{jensen2018knowing}. Further findings suggest that people feel comfortable with drone's speed being 0.5 m/s  \cite{wojciechowska2019collocated} and 0.7 m/s \cite{jensen2018knowing}, altitude being 1.2 m \cite{yeh2017exploring}, lateral distance being 0.6 m \cite{yeh2017exploring}, with the frontal side as a direction of approach, and with a direct trajectory \cite{wojciechowska2019collocated}. For the guiding outdoors scenario, the preferred distance to the robot and height were 4.0 m and 2.6 m respectively \cite{kong2018effects}. Proxemics with the drones varied based on gender as female and male participants preferred distances were 1.5 m and 1.1 m respectively \cite{yeh2017exploring}. Cultural differences might play a role in these preferences as participants in China were comfortable with letting the drone into their intimate zone \cite{jane2017drone}. 

Several works designed and evaluated communication interfaces for UAVs: high-level signal communication mechanisms (blinker, beacon, thruster, and gaze) to communicate drone's intentions \cite{szafir2014communication}, four gestures (waggle, nod, toss, and orienting) to signal acknowledgement to an interacting human partner \cite{jensen2018knowing}, and an LED visual feedback for the blimp to respond to human gestures \cite{yao2019autonomous}. The majority of them were positively received by end users. Additionally, animal-inspired behavioral patterns were also explored: the low speed of 2 m/s as well as high cyclicity (i.e. repetition) expressions can be used to increase distancing in human-drone interactions \cite{duncan2017effects}. 

Likewise, other works focused on the analysis of how humans could control and interact with drones: voice and gestures (e.g. fly closer, follow, take a selfie, etc.) \cite{cauchard2015drone}, and hand waving \cite{yao2019autonomous}. 

A number of works focused on improving drone's standard designs by making the drones safe-to-touch \cite{abtahi2017drone}, social \cite{yeh2017exploring}, and home companions \cite{karjalainen2017social}. Such design decisions successfully increased perceived safety: safe-to-touch drones were allowed into people's intimate zone \cite{abtahi2017drone} while a social drone was allowed to come 30\% closer than a non-social drone \cite{yeh2017exploring}.   

Drone's movements, distance, speed, height (altitude) and trajectory were among the affecting factors concerning perceived safety \cite{chang2017spiders}. Design factors such as size being too large or too small, squared shape, dark colors often influence perceived safety negatively \cite{cauchard2015drone, jones2016elevating, yeh2017exploring, chang2017spiders}. Propeller noise and produced wind created additional discomfort \cite{cauchard2015drone, jones2016elevating, yeh2017exploring, chang2017spiders}. Furthermore, sudden  speeding  up,  unstable  maneuvers  and unusual  movements  such  as  flips  and  back-and-forth  movements  were  perceived  as  threatening, aggressive, erratic and  potentially  dangerous \cite{chang2017spiders, jensen2018knowing}. 

In general, people feel safe around medium-sized drones \cite{duncan2013comfortable, yeh2017exploring} regardless of their level of autonomy \cite{kong2018effects, wojciechowska2019collocated}. Further findings suggest that drones' steady movements and predictable trajectory, production of social gestures, the presence of safe guards, face and feedback lights can increase people's feelings of safety around drones \cite{abtahi2017drone, karjalainen2017social, chang2017spiders, jensen2018knowing, yeh2017exploring}. 

Many authors suggest that design decisions should inform drones' purpose as it can guide people in their expectations and interactions with the drones \cite{chang2017spiders, jensen2018knowing}. A safe or a threatening drone can be designed using colors, logos and decorations, a balanced size and shape, and with the help of carefully engineered movements and gestures. Additionally, it is advised to communicate safety messages related to drone's malfunctioning, taking off and landing, battery life, and close proximity to obstacles \cite{jones2016elevating}. Moreover, further concerns are related to drones being near buildings,  other drones, and wildlife \cite{chang2017spiders}. 

Numerous studies found supporting evidence that people have a tendency to animate drones and compare them to pets \cite{cauchard2015drone, abtahi2017drone, wojciechowska2019collocated}. However, it might not be the case for every culture as participants in China were less likely to compare drones to pets \cite{jane2017drone}.

\section{Autonomous vehicles \cite{waytz2014mind, gold2015trust, hauslschmid2017supportingtrust, forster2017increasing, rothenbucher2016ghost, reig2018field, jayaraman2019pedestrian, clamann2017evaluation, palmeiro2018interaction, ackermann2019experimental, dey2019pedestrian, eden2017expectation, nordhoff2020passenger, rahmati2019influence, paddeu2020passenger}} \label{sec:AVs}

\subsection{Overview of works on autonomous vehicles}
The articles presented in this section focus on autonomous vehicles (AVs), i.e., vehicles capable of sensing their environment and operating without human involvement. By introducing AVs into people's daily lives, positive impacts, for instance on road safety or fuel consumption, are expected \cite{gold2015trust}. However, one of the required conditions for the successful integration of AVs is their acceptance by various road users. Therefore, research works investigating perceived safety of AVs can be categorized based on who the road user was: a driver that has to give away control \cite{waytz2014mind, gold2015trust, forster2017increasing, hauslschmid2017supportingtrust}, a pedestrian that has to make road-crossing decisions \cite{rothenbucher2016ghost, reig2018field, clamann2017evaluation, palmeiro2018interaction,  ackermann2019experimental, dey2019pedestrian, jayaraman2019pedestrian}, a passenger of a shared public transport \cite{eden2017expectation, nordhoff2020passenger, paddeu2020passenger} or the driver of another vehicle \cite{rahmati2019influence}.  

Although the validity of the experiments is higher when real AVs were used (such as Uber AV \cite{reig2018field}, Smart Shuttle \cite{eden2017expectation}, the automated shuttle ``Emily'' from Easymile \cite{nordhoff2020passenger}, Shared Autonomous Vehicle shuttle \cite{paddeu2020passenger}, and Chevy Bolt \cite{rahmati2019influence}), several works utilized driving simulations (such as a National Advanced Driving Simulator \cite{waytz2014mind}, a BMW Series-6 simulator \cite{gold2015trust}, a motion-based driving simulator at the Wuerzburg Institute for Traffic Sciences (WIVW GmbH) \cite{forster2017increasing}), videos of real cars \cite{hauslschmid2017supportingtrust, ackermann2019experimental, dey2019pedestrian}, virtual environment \cite{jayaraman2019pedestrian}, and ordinary cars controlled in a Wizard-of-Oz fashion  \cite{rothenbucher2016ghost, clamann2017evaluation, palmeiro2018interaction}. 

\subsection{Selected works on autonomous vehicles}
One of the earliest studies was conducted in 2014 by Waytz et al. \cite{waytz2014mind} with 100 participants that used a National Advanced Driving Simulator, where people drove either a normal car, an autonomous vehicle (a vehicle capable of controlling its steering and speed), or a comparable autonomous vehicle that had a name, gender and voice. Their findings suggest that trust was predicted based on people's tendency to anthropomorphise: people trusted the anthropomorphic vehicle more than those in the autonomous condition, who in turn trusted their vehicle more than those in the normal condition. It was also found that anthropomorphism affects attributions of responsibility or punishment as people tended to blame the car for the simulated accident. 

Another study on trust in automation was investigated by Gold et al.  \cite{gold2015trust} who conducted a BMW Series-6 simulator study with 72 participants who drove on a highway at a speed of 120 km/h and experienced three take-over scenarios for which they had to regain control and brake or steer to change lanes and avoid the collision. Results suggest that such driving experience increased self-reported trust in automation while leading to a decrease in driver discharge and safety gain. Participants over 60 years old demonstrated a more positive rating of the system and higher trust compared to people younger than 30.

Hauslschmid et al. \cite{hauslschmid2017supportingtrust} investigated how trust can be increased by means of a driver interface that visualizes the car's interpretation of the current situation and its corresponding actions. To this end, 30 participants watched the videos of real driving inside the real car test setup and compared three types of visualizations overlaid to a driving scene: a chauffeur avatar, a world in miniature, and a display of the car's indicators (left, right arrows) as the baseline. The world in miniature was the most effective visualization in increasing trust. It also fostered the strongest feeling of safety as well as the best user experience. 

Forster et al. \cite{forster2017increasing} conducted a study with 17 drivers in the motion-based driving
simulator at the Wuerzburg Institute for Traffic Sciences
(WIVW GmbH) which confirmed that semantic speech output of the AV's actions increased the level of trust, anthropomorphism, and acceptance of AVs. 

One of the earliest studies that investigated pedestrians' perception of AVs was an observational field study by Rothenburcher et al. in 2016 \cite{rothenbucher2016ghost}. They designed a car seat costume that concealed the driver for the vehicle to appear to have no driver. Their observations and interviews with 67 pedestrians that encountered that ``self-driving" vehicle at the crosswalk and reported seeing no driver suggest that pedestrians generally adhered to existing interaction patterns with cars except when the car misbehaved by moving into the crosswalk which caused some degree of hesitation. 

Reig et al. \cite{reig2018field} interviewed 32 pedestrians after their interaction with the Uber AV in a field study in 2018. Their findings revealed that there was an inherent relationship between favorable perceptions of technology and feelings of trust towards AVs, which was in turn influenced by a favorable interpretation of the company's brand. 

Several works designed vehicle-to-pedestrian communication displays for AVs and evaluated pedestrians' levels of perceived safety while making road-crossing decisions. Clamann et al.  \cite{clamann2017evaluation} conducted a Wizard-of-Oz study with 55 participants that evaluated messages displayed on a 32-inch (81 cm) LCD mounted on the front of the Dodge Sprinter van that was reported to participants as an autonomous vehicle. Analysis of ``safe" vs ``unsafe" crossing decisions made by the participating pedestrians revealed that there were no significant differences between display conditions (advice, information, off, and no display). When asked to identify the most important information needed to cross safely in front of the AV, participants rated that distance to the vehicle (56\%) was the most important factor in their decision to cross followed by the speed (46\%) and traffic density (24\%). Despite these results, nearly half of the participants (46\%) also reported that having displays like the ones used in the experiment would be helpful when autonomous vehicles become available. 

Similar findings were obtained in the Wizard-of-Oz experiment by Palmero et al. \cite{palmeiro2018interaction} with 24 pedestrians making crossing decisions. While there were also no significant differences between vehicle conditions (traditional vehicle, driver reading a newspaper, inattentive driver in a vehicle with a ``self-driving" sign on the roof, inattentive driver in a vehicle with ``self-driving" signs on the hood and door, attentive driver), vehicle behavior (stopping vs. not stopping), and approach direction (left vs. right), distance and speed were more important for pedestrians in deciding to cross the road than were features related to AVs (e.g., external messages on the vehicle). Video-based and GPS-based measurements of the critical gap distance were calculated and ranged between 5.66 and 7.67~s. 

Ackermann et al. \cite{ackermann2019experimental} conducted a study with 26 participants who were presented with 20 augmented variations of a real-world video of an AV that was recorded from a pedestrian perspective and dynamically augmented to create realistic 20 variations differing in technology (LED display, Projection, or LED light strip), position, coding (text or symbol), content of the message, and display mode. Their findings suggest that projections and pedestrian advice (``Go ahead") are more comfortable than displays and vehicle status, independent of text or symbol-based presentation. 

Dey et al. \cite{dey2019pedestrian} investigated pedestrians' road-crossing behavior in a video-based experiment. Their findings suggest that participants based their decisions on vehicle's behavior such as maintaining speed or slowing down regardless of the driving mode (manually-driven vs automated), vehicle appearance and size (futuristic vs ordinary), and distances (45 m to 1.5 m). However, when the intent of the vehicle was not clear, there were differences between the two vehicles at certain distances: a futuristic-looking car was trusted less than an ordinary-looking vehicle. 

In a later study with 30 participants in a virtual reality environment by Jayaraman et al. \cite{jayaraman2019pedestrian}, pedestrians trusted the AVs more when the AVs exhibited defensive driving behavior and when crosswalks had traffic signals unaffected by the driving behavior of the AVs (defensive, normal, and aggressive). There were also strong correlations between trust in AVs and pedestrians' gaze, distance to the vehicle, and jaywalking time.

There have also been studies which explored how safe passengers would feel sharing AVs in the domain of public transportation. The Smart Shuttle was used for a field study with 17 participants in the work by Eden et al. \cite{eden2017expectation}. Although some participants expressed safety concerns prior to using the Smart Shuttle, all of them were cleared after the ride as it was operating at a low speed (20 km/h) and in the presence of researchers. Participants commented on the absence of the seatbelts as the sudden hard stops made people lurch forward. Most participants agreed that their perceived safety might have decreased if it they had been on a large-sized bus with no attendants on board, travelling on an actual route at a regular speed.

Nordhoff et al. \cite{nordhoff2020passenger} conducted a real-world study with 119 participants who rode the automated shuttle `Emily' from Easymile (2nd generation EZ10) and associated their perception of safety with the low speed, the ability of the automated shuttle to identify and react to other road users and traffic objects in the external environment, the smooth and passive longitudinal and lateral vehicle control, the possibility of pressing the emergency button, and their general trust in technology.

Paddeu et al. \cite{paddeu2020passenger} conducted a study where two unrelated participants experienced riding in a Shared Autonomous Vehicle (SAV) shuttle for a total of 55 participants. Trust was significantly affected by each independent variable: direction of face (forwards and backwards) and maximum vehicle speed (8 and 16 km/h). Additionally, trust had a strong correlation with perceived comfort. 

With the aim to understand other drivers' behaviors when sharing a road with AVs, Rahmati et al. \cite{rahmati2019influence} conducted a study with 9 drivers who engaged in a car-following task within a platoon of three vehicles, where the second vehicle was either a human-driven car or an AV (Texas A\&M University's automated Chevy Bolt). A data-driven and a model-based approach were used to identify possible changes in the participants' driving behavior between the two driving mode conditions. Their findings suggest that participating drivers of the third vehicle indeed had varying driving behaviors in these two conditions: human drivers felt more comfortable following the AV as they drove closer to AV and put less weight on the crash risk.

\begin{table*}[tbp]
\label{tab:focus}
\centering
\begin{tabular}{|p{0.135\linewidth}|p{0.104\linewidth}|p{0.104\linewidth}|p{0.104\linewidth}|p{0.104\linewidth}|p{0.134\linewidth}|p{0.134\linewidth}|}
\hline
                   & \textbf{Industrial\newline manipulators} & \textbf{Mobile\newline robots} & \textbf{Mobile\newline manipulators} & \textbf{Humanoid robots} & \textbf{Drones} & \textbf{Autonomous\newline vehicles}  \\
\hline\hline
\centering{\textbf{Distance}} & \cite{arai2010assessment} \cite{weistroffer2014assessing} \cite{hocherl2017motion} \cite{you2018enhancing} \cite{bergman2019close} & \cite{koay2006empirical} \cite{huttenrauch2006investigating} \cite{walters2008human}  & \cite{macarthur2017human} & \cite{kanda2002development} & \cite{cauchard2015drone} \cite{yeh2017exploring}  \cite{acharya2017investigation} \cite{chang2017spiders} \cite{jane2017drone}  \cite{kong2018effects} \cite{wojciechowska2019collocated}   & -\\
\hline
\centering{\textbf{Robot speed}}  & \cite{kulic2005anxiety} \cite{kulic2007physiological} \cite{zoghbi2009evaluation} \cite{arai2010assessment} \cite{ng2012impact} \cite{zhao2020task} & - & \cite{dehais2011physiological} \cite{strabala2013toward} \cite{macarthur2017human} & - & \cite{chang2017spiders} \cite{jensen2018knowing}  \cite{wojciechowska2019collocated}    & -\\
\hline
\centering{\textbf{Robot speed\\ $\propto$ distance}} & \cite{yamada1999proposal} \cite{hanajima2005motion} \cite{lasota2015toward} & - & \cite{brandl2016human} & -  & - & \cite{rothenbucher2016ghost} \cite{clamann2017evaluation} \cite{dey2019pedestrian} \cite{jayaraman2019pedestrian} \\
\hline
\centering{\textbf{Direction of approach}} & - & \cite{koay2005methodological} \cite{woods2006methodological} \cite{syrdal2006doing} \cite{dautenhahn2006may} \cite{walters2007robotic} \cite{karreman2014robot}  & - & \cite{koay2009five} &  \cite{wojciechowska2019collocated} & -\\
\hline
\centering{\textbf{Robot size \& appearance}} & \cite{charalambous2016development} & \cite{bhavnani2020attitudes} & - & \cite{koay2007living} \cite{kanda2008analysis} \cite{rajamohan2019factors} & \cite{duncan2013comfortable} \cite{cauchard2015drone} \cite{jones2016elevating} \cite{yeh2017exploring} \cite{chang2017spiders} & \cite{dey2019pedestrian} \\
\hline
\centering{\textbf{Motion fluency \& predictability}} & \cite{charalambous2016development} \cite{hocherl2017motion} \cite{bergman2019close} \cite{koert2019learning} \cite{aeraiz2020robot} \cite{zhao2020task} & \cite{syrdal2009negative} & \cite{dehais2011physiological} \cite{strabala2013toward} \cite{dragan2015effects} & \cite{huber2008human} \cite{koay2009five} & \cite{chang2017spiders} \cite{jensen2018knowing}  & \cite{rothenbucher2016ghost} \cite{clamann2017evaluation} \cite{dey2019pedestrian} \cite{jayaraman2019pedestrian}\\
\hline
\centering{\textbf{Communication}} & \cite{arai2010assessment} \cite{koert2019learning} & - & - & \cite{moon2014meet} \cite{rajamohan2019factors} & \cite{szafir2014communication} \cite{jones2016elevating} \cite{jensen2018knowing} \cite{yao2019autonomous}  &  \cite{waytz2014mind} \cite{clamann2017evaluation} \cite{forster2017increasing} \cite{hauslschmid2017supportingtrust}  \cite{palmeiro2018interaction} \cite{ackermann2019experimental}\\
\hline
\end{tabular}
\caption{Factors determining perceived safety. In each row, a different factor is listed: distance, robot speed, proportionality between robot speed and distance, direction of approach, robot size and appearence, motion fluency and predictability, communication, smooth contacts. In each column, a different robot type is considered, and precisely industrial manipulators, mobile robots, mobile manipulators, humanoid robots, drones, and autonomous vehicles.}
\end{table*}

\subsection{Achieving perceived safety for autonomous vehicles}
The analysis of the studied articles showed that equipping the AV with a driver interface that either visualizes the car's interpretations and actions (e.g. the world in miniature \cite{hauslschmid2017supportingtrust}) or has additional anthropomorphic features such as a name, gender and voice \cite{waytz2014mind, forster2017increasing}, increases the level of perceived safety.
Interestingly, participants over 60 demonstrated a more positive rating of the system and higher trust compared to people below 30 years of age \cite{gold2015trust}. 

From the analysis of the studies investigating pedestrians' road-crossing decisions in front of AVs, most findings confirm that individual predispositions and legacy behaviors with ordinary vehicles such as distance and speed are more important for pedestrians in deciding to cross the road as reported in \cite{rothenbucher2016ghost, clamann2017evaluation, dey2019pedestrian} than are features related to AVs (e.g., external communication displays or signs on the vehicle \cite{clamann2017evaluation, palmeiro2018interaction}, AVs' driving mode (manually-driven vs automated) \cite{ackermann2019experimental, dey2019pedestrian}, vehicle appearance and size (futuristic vs ordinary) \cite{dey2019pedestrian}). Despite insignificant differences found between various types of vehicle-to-pedestrian communication interfaces, many pedestrians reported that having pedestrian advice would be helpful when AVs become available independently of message content or display mode used \cite{clamann2017evaluation, ackermann2019experimental}. Additionally, pedestrians based their road-crossing decisions on AV's driving behavior (such as defensive driving, maintaining speed, and slowing down) \cite{dey2019pedestrian, jayaraman2019pedestrian}. It is important to note that pedestrians take into account other factors, such as non-verbal communication with the driver (e.g. eye contact) or presence of traffic signals \cite{jayaraman2019pedestrian}.

Passengers' safety concerns were cleared after experiencing the actual ride in a shared AV as participants associated perceived safety with low speed (8, 16 and 20 km/h), the ability of the automated shuttle to identify and react to other road users and traffic objects in the external environment, the smooth and passive longitudinal  and  lateral  vehicle  control,  presence of the  emergency button inside the shuttle, forward facing and their general trust in technology \cite{eden2017expectation, nordhoff2020passenger, rahmati2019influence}. However, most participants agreed that their perceived safety might decrease if it was a large-sized bus with no attendants on board travelling on an actual route at a regular speed.

Human drivers of vehicles felt comfortable following the AV as they drove close to AV and put less weight on the crash risk \cite{paddeu2020passenger}. 

Generally, there was an inherent relationship between favorable perceptions of technology and feelings of trust towards AVs, which was in turn influenced by a favorable interpretation of the company's brand (e.g. Uber, BMW, and others) \cite{reig2018field}. 

\begin{figure*}[tbp]
	\centering
	\includegraphics[width=\textwidth]{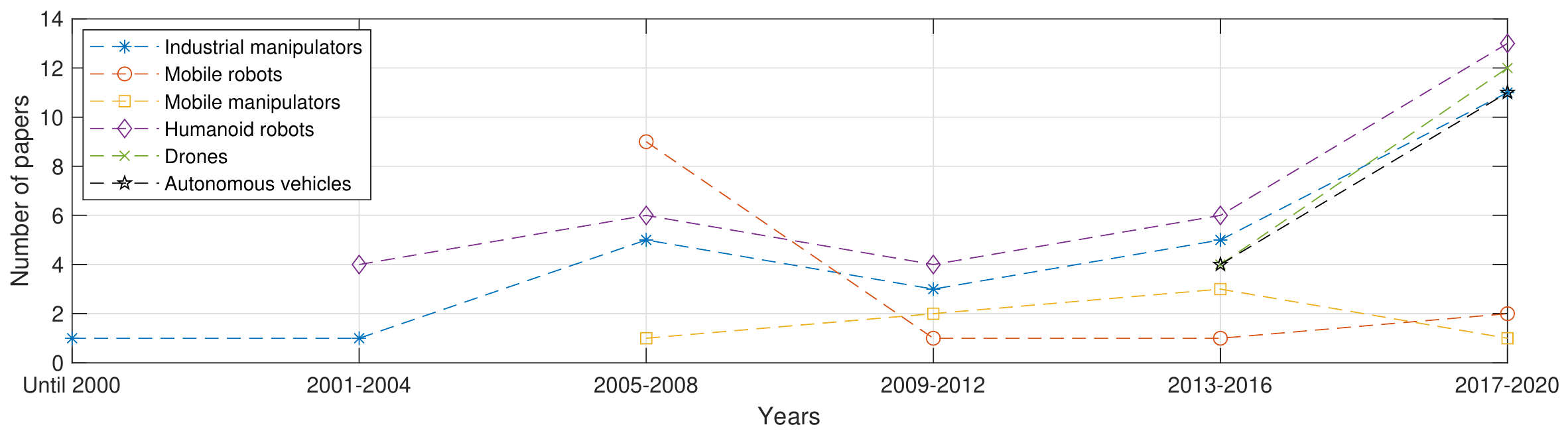}
	\caption{Time evolution of quantity of published works (analyzed in this survey) for each robot type.}
	\label{fig:years}
\end{figure*}

\section{Discussion}\label{sec:disc}
\subsection{Factors determining perceived safety}
Table 4 summarizes the main factors related to the robot motion and characteristics that determine perceived safety across all robot types analyzed in the Sections \ref{sec:IM}-\ref{sec:AVs}.
\begin{enumerate}
    \item Distance: the largest the distance between human and robot, the safest the robot is perceived. In some cases, distances were related with the different zones of Hall's proxemics.
    \item Robot speed: robots moving with lower speeds are perceived as safer.
    \item Robot speed $\propto$ distance: higher and higher speeds are considered safe if the robot is farther and farther from the human. This is in line with standards for actual safety (specifically, speed and separation monitoring \cite{haddadin2016physical}), as this allows the robot to stop before a possible contact with the human occurs. For AVs, we interpret the indicator to refer to the fact that a car is perceived as safe if it slows down as it approaches participants (e.g., for pedestrian crossing).
    \item Direction of approach: certain directions of approach (e.g., from the front-side rather than directly from the front, or from the back) are perceived as safer.
    \item Robot size \& appearance: robots that are larger or with certain features (e.g., shape and color) are perceived as less safe, regardless of their motion.
    \item Motion fluency \& predictability: a fluent and predictable motion of the robot leads to a better perception of safety.
    \item Communication: adding cues that hint at the specific robot motion (e.g., emitting a sound before the robot starts moving) improves perceived safety.
    \item Smooth contacts: in case of contacts (e.g., during handover), the absence of abrupt robot motions positively contributes to perceived safety.
\end{enumerate}

It is important to notice that certain aspects are ubiquitous (e.g., motion fluency and predictability), while others apply only to certain types of robots (e.g., exchanging smooth contact forces is applicable for an handover task with an industrial manipulator, but not for an interaction with a drone). Furthermore, there are aspects that are neglected for certain types of robots, only because of the nature of the experiment: for instance, the robot speed was never considered a factor related to perceived safety in papers on mobile robots because the robot speed was always very low in the related experiments. 

\subsection{Experiment duration and location}
Short-term experiments which lasted from minutes to hours, typically conducted within the same day for each participant, were in general preferred. The reason is that this experiment duration was enough to allow researchers to study people's reactions, as well as to test new assessment methods. The exception was constituted by studies such as \cite{hanajima2006further, koay2007living, koay2009five}, which ran longer experiments, within a time span of up to two months, in order to assess the effects of long-term habituation of the human subjects to the interaction with the robot.

Experiments were conducted in (i) research laboratories, (ii) factories, (iii) locations close to the living environments of the participants, (iv) fairs and symposia and, finally, (v) virtual environments and driving simulators.

Most studies were conducted in laboratory settings, due to the obvious advantages in terms of setting up the experiments. The experiments conducted in factories were aimed at assessing the level of safe interaction between humans and robots in conditions closer to those of manufacturing environments. The experiments conducted in locations closer to real living conditions were run in houses, apartments, rooms, or outdoors close to buildings and roads (for drones and autonomous vehicles) so as to increase the level of natural human-robot interaction. The participants noticed that, due to these conditions, they did not feel as if they were being tested and could behave more naturally. Thus, based on these articles, we can conclude that this type of approach can increase the level of truthfulness of participants' reactions.
In two papers, experiments were conducted during trade fairs \cite{maurtua2017human} and in the symposium reception event \cite{walters2008human}, in order to have easy access to participants with different levels of experience with robotics
technologies. Some papers used virtual reality and driving simulators: the reason for integrating this technology into the experiments was typically to avoid safety risks, while still providing an immersive experience to the participants. In general, as stated by Weistoffer et al., ``virtual reality may be a good tool to assess the acceptability of human-robot collaboration and draw preliminary results through questionnaires, but that physical experiments are still necessary to a complete study, especially when dealing with physiological measures''  \cite{weistroffer2014assessing}.

\subsection{Trends over the years}
In Fig. \ref{fig:years}, we report the time evolution of the number of published papers, among those analyzed in our survey, for each robot type. One can see that industrial manipulators were the first type of robots to be considered, followed by humanoids. It is possible to observe a surge of interest in the years 2005-2008 on mobile robots, while the number of papers on industrial manipulators, humanoid robots, drones and autonomous vehicles show a sharp increase in the last years: this further justifies the need for our review, as more and more works will likely be published on perceived safety in the coming years.

\section{Conclusions}\label{sec:concl}
This survey paper has highlighted the lack of uniform terminology for perceived safety in pHRI: one of the purposes of this work was to provide an overview of how these terms are used in the literature, and to easily allow researchers to locate the related papers. Regarding assessment methods, we replicated a standard categorization, analyzing how these methods are employed across the six considered robot types. More than one hundred papers were analyzed, highlighting common trends and themes in terms of target users, experimental conditions, and connection between robot characteristics and motion on the one hand, and perceived safety on the other hand. 

\section*{Acknowledgements}
This work was supported by Nazarbayev University under Collaborative Research Project no. 091019CRP2118.


\balance
  \bibliographystyle{elsarticle-num} 
  \bibliography{bibliography}


\end{document}